\documentclass[conference,table,letterpaper]{ieeeconf}
\IEEEoverridecommandlockouts
\usepackage[algoruled,vlined,linesnumbered]{algorithm2e}
\usepackage{amsmath,amssymb,latexsym,float,epsfig,subfigure}

\usepackage[dvipsnames]{xcolor}
\usepackage{tikz}

\usepackage{tikz-3dplot}
\usepackage{pgfplots}
\usepackage{smartdiagram}
\usepackage{url}
\usetikzlibrary{shapes,arrows,calc}
\usepackage{tkz-euclide}
\usepackage{adjustbox}
\usepackage{bm}
\usepackage{tcolorbox}
\usepackage{booktabs}
\usepackage{colortbl}
\usepackage[]{caption}
\usepackage{multirow}
\usepackage{listings}
\usetkzobj{all}
\usepackage{cite}

\newcommand{\shrinka}{\def\baselinestretch{0.95}\large\normalsize}
\SetCommentSty{mycommfont}
\newcommand{\ArmMetricSet}{$\mathcal{M}_{A}$}
\newcommand{\GraspMetricSet}{$\mathcal{M}_{G}$}

\newcommand{\GraspSet}[1]{$\mathcal{G}_{#1}$}
\newcommand{\GoalSet}[1]{$\mathcal{W}_{#1}$}
\newcommand{\grasp}[1]{\ensuremath{\mathbf{g}_{#1}}}

\newcommand{\Tf}[2]{$^{#1}T_{#2}$}
\newcommand{\Pos}[2]{$^{#1}P_{#2}$}

\newcommand{\Object}[1]{$\mathcal{O_{#1}}$}

\newenvironment{itemize_tight}{
\begin{itemize}
  \setlength{\itemsep}{1pt}
  \setlength{\parskip}{0pt}
  \setlength{\parsep}{0pt}
}{\end{itemize}}

\newenvironment{enumerate_tight}{
\begin{enumerate}
  \setlength{\itemsep}{1pt}
  \setlength{\parskip}{0pt}
  \setlength{\parsep}{0pt}
}{\end{enumerate}}

\def\sref#1{Section~\ref{#1}}
\def\aref#1{Algorithm~\ref{#1}}
\def\fref#1{Figure~\ref{#1}}
\def\tref#1{Table~\ref{#1}}

\usetikzlibrary{mindmap,trees}

\definecolor{bgcolor}{rgb}{0.98, 0.98, 1.0}
\definecolor{bgcolorGray}{rgb}{0.8, 0.8, 0.8}
\definecolor{lightGray}{rgb}{0.9,0.9,0.9}
\tikzset{
  basic/.style = {draw,text width=2cm,drop shadow,rectangle},
  root/.style = {basic,rounded corners=2pt,thin,align=center,fill=gray!50},
  level 2/.style = {basic,rounded corners=6pt,thin,align=center,fill=gray!25,
    text width=6em},
  level 3/.style={basic,thin,align=center,fill=pink!50,text width=4.5em,sibling distance=2em}
}
\colorlet{tableheadcolor}{gray!25}
\newcommand{\headcol}{\rowcolor{tableheadcolor}}
\colorlet{tablerowcolor}{gray!10}

\newcommand{\topline}{\arrayrulecolor{black}
  \specialrule{0.1em}{\abovetopsep}{0pt}%
  \arrayrulecolor{tableheadcolor}\specialrule{\belowrulesep}{0pt}{0pt}%
  \arrayrulecolor{black}}
\newcommand{\midline}{\arrayrulecolor{tableheadcolor}
  \specialrule{\aboverulesep}{0pt}{0pt}%
  \arrayrulecolor{black}\specialrule{\lightrulewidth}{0pt}{0pt}%
  \arrayrulecolor{white}\specialrule{\belowrulesep}{0pt}{0pt}%
  \arrayrulecolor{black}}
\begin{document}
\shrinka

%%%%%%%%%%%%%%%%%
%% Paper Info
%%%%%%%%%%%%%%%%%
\title{Grasp selection analysis for two-step manipulation tasks}
\author{Ana Huam\'{a}n Quispe$\quad$ 
\thanks{Institute for Robotics and Intelligent Machines, Georgia Institute of Technology, Atlanta, GA 30332, USA. {\tt\small ahuaman3@gatech.edu}}}
\maketitle

%%%%%%%%%%%%%%%
%% Abstract
%%%%%%%%%%%%%%%
\begin{abstract}
Manipulation tasks are sequential in nature. Grasp selection approaches
that take into account the constraints at each task step are critical, since they allow
to both (1) Identify grasps that likely require simple arm motions through the whole task and
(2) To discard grasps that, although feasible to achieve
at earlier steps, might not be executable at later stages due to goal task constraints. In this
paper, we study how to use our previously proposed manipulation metric for tasks in which 2 steps
are required (pick-and-place and pouring tasks). Even for such simple tasks,
it was not clear how to use the results of applying our metric (or any metric for that matter) to rank all the candidate grasps: Should only the start state be considered, or only the goal, or a combination of both? In order to find an answer, we evaluated the (best) grasps selected by our metric under each of these
3 considerations. Our main conclusion was
that for tasks in which the goal state is more constrained (pick-and-place), using a combination of the metric measured at the start and goal states renders better performance when compared with choosing any other candidate grasp, whereas in tasks in which the goal constraints are less rigidly defined, the metric measured at the start state should be mainly considered. We present quantitative results in simulation and validate our approach's practicality with experimental results in our physical robot manipulator, \textit{Crichton}.
%evaluated only at the start step. We present quantitative results in simulation and validate our approach's practicality with experimental results in our physical robot platform.
%We evaluate our metric's ability to select grasps under 3 modalities: Evaluation at the start step, at the goal step and an average  of both. We conclude that for tasks in which the goal state is more constrained (pick-and-place), the metric as an average renders better performance, whereas in tasks in which the goal constraints are loose, we obtain better results using the metric
%evaluated only at the start step. We present quantitative results in simulation and validate our approach's practicality with experimental results in our physical robot platform.
\end{abstract}

%\IEEEpeerreviewmaketitle % Mm. What is this? :D

%%%%%%%%%%%%%%%%%%
%% Introduction
%%%%%%%%%%%%%%%%%%
\section{Introduction}
\label{sec:Introduction}
Given a manipulation task, a redundant robot manipulator and a target object, many possible candidate grasps can be used
to accomplish the task. Finding a suitable grasp among the infinite set of candidates
is a challenging problem that has been addressed frequently in robotics, resulting in an
abundance of approaches~\cite{bohg2014data}. Interestingly, the vast majority of these methods have two aspects in common: (1) The metrics used for grasp selection focus on the
hand-centric aspect of a manipulation task, such as grasp robustness, and
%measured with either analytical \cite{ferrari1992planning} or heuristic measures \cite{balasubramanian2012physical}. 
(2) Manipulation is implicitly seen as a single-step 
task, in which the main goal is to reach an object without further regard to what will be done
with it once it is grasped. 

In general, even the simplest of manipulation tasks, such as pick-and-place, has 2 or more steps. And while grasp robustness is perhaps the most important aspect to predict the success of a manipulation task,
it is not the only factor to consider. For a grasp to be executable through a whole task, feasible arm motions between steps are needed. We argue that a metric that considers both the grasp robustness and arm kinematics 
%(dependent on the task constraints) 
is a more useful way to select grasps that will in turn
entail arm motions that can be easily planned and executed in the real world.
%, producing shorter end-effector displacements when compared to motions generated by selecting other (perhaps harder to reach) candidate grasps.

In our previous work \cite{huaman2016metric}, we proposed our arm-and-grasp metric~($m_{ag}$) and
showed its usefulness to select grasps for simple pick-up tasks that in average involved shorter
and faster arm motions when  compared with those produced by selecting other possible candidate grasps. In this paper we extend the use of $m_{ag}$ to select grasps for tasks with 
2 steps, such as pick-and-place and pouring tasks. A challenge in considering 2-step vs 1-step 
tasks is the fact that the way to use $m_{ag}$ is not straightforward: Should the metric be measured
only at the start state, at the goal stage or should it be a combination of measurements at both states? Intuitively, we suspect that a metric
that considers the whole task (from start to goal) would be more useful; however, after analyzing
the 2 tasks aforementioned, we conclude that considering either start, goal or both mainly
depends on how constrained/loose the goal state is. We will show simulation experiments that
support this claim in \sref{sec:PickAndPlaceTasks} and \sref{sec:PouringTasks}. The remainder of this paper is organized as follows: Section \ref{sec:RelatedWork} 
presents a condensed review of existing work. In section \ref{sec:RecapMetric} we present a brief summary of the arm-and-grasp metric $m_{ag}$ already presented in \cite{huaman2016metric}. Section \ref{sec:PickAndPlaceTasks} presents our evaluation of the metric $m_{ag}$ in pick-and-place tasks mesured at the start and goal state as well as an average of both. Section \ref{sec:PouringTasks} shows a similar evaluation but in pouring tasks. Finally, our conclusions, including the limitations of our approach are briefly stated in Section \ref{sec:Conclusions}.

%-------------------------
%Cover Image
%-------------------------
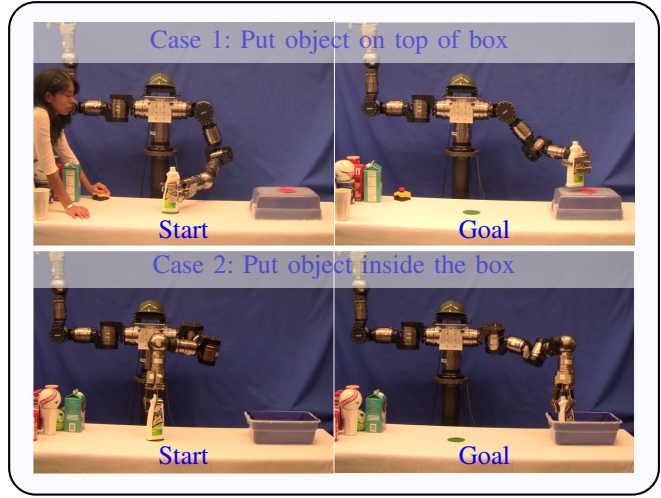
\begin{figure}[t]
\centering
\begin{tikzpicture}[framed,background rectangle/.style={thick,draw=black,fill=bgcolor,rounded corners=1em}]
\node[]{\includegraphics[width=0.45\textwidth]{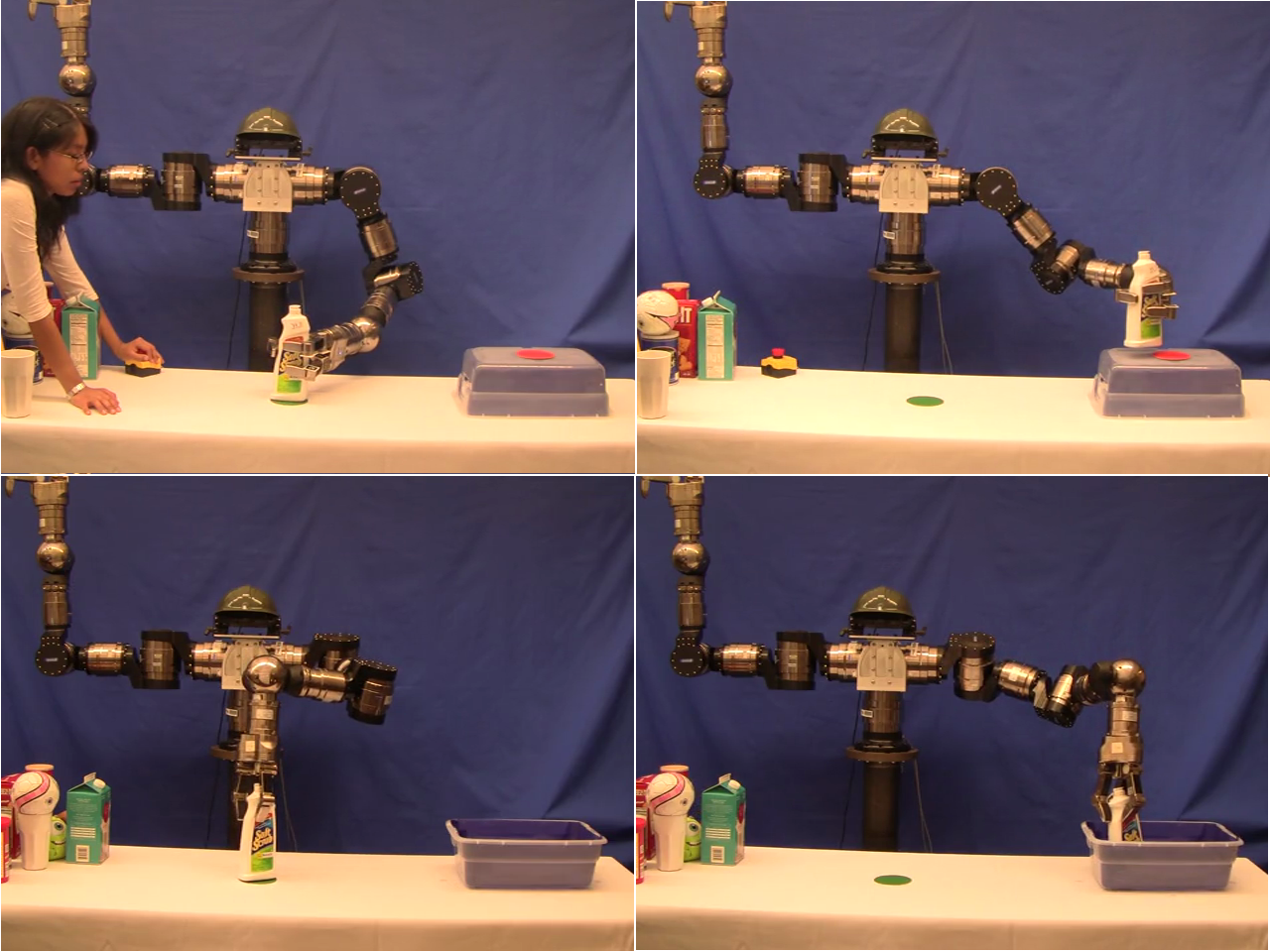}};
\draw[line width=0.75mm, white] (-4,0) -- (4,0);
\node[text=blue,xshift=0cm,yshift=2.75cm,fill=white,fill opacity=0.5]{~~~~~~~~~~~~~Case 1: Put object on top of box~~~~~~~~~~~~~~~};
\node[text=blue,xshift=-2.0cm,yshift=0.25cm]{Start};
\node[text=blue,xshift=2.0cm,yshift=0.25cm]{Goal};
\node[text=blue,xshift=0cm,yshift=-0.25cm,fill=white,fill opacity=0.5]{~~~~~~~~~~~~~Case 2: Put object inside the box~~~~~~~~~~~~~~};
\node[text=blue,xshift=-2.0cm,yshift=-2.75cm]{Start};
\node[text=blue,xshift=2.0cm,yshift=-2.75cm]{Goal};
\end{tikzpicture}
\caption{Pick-and-place sample tasks: In Case 1, our metric $m_{ag}$ averaged
over the start and goal state deemed the side-grasp shown as best for the task at hand. In case 2, the top-grasp shown is now considered the best option given the collision scenario (the side-grasp is not even considered given its infeasibility to be executed due to the goal constraints).}
\label{fig:coverImage}
\vspace{-2.0em}
\end{figure}

%%%%%%%%%%%%%%%%%%
%% Previous work
%%%%%%%%%%%%%%%%%%
\section{Related Work}
\label{sec:RelatedWork}
In this section we review work concerning grasp selection for manipulation tasks. 
For a more detailed survey of previous research in the area, the reviews presented in
\cite{bohg2014data} and \cite{sahbani2012overview} are highly recommended. 

Pioneering work on grasp selection was developed by Cutkosky \cite{cutkosky1989grasp}, who
observed that humans select grasps in order to satisfy 3 main types of constraints:
Hand constraints, object constraints and task-based constraints. As pointed out by Bohg et al. in \cite{bohg2014data}, there is little work on task-dependent grasping when compared to work focused on the first two types of constraints. Hence, the main goal for most existing planners is to find a grasp such that the robot can reach the target object, without further regard of what will be done once the object is picked.

In current common practice, grasps are generated offline and are then ranked based on their force-closure properties, which theoretically express their 
robustness and stability. One of the most popular metrics ($\epsilon$) was proposed
 by Ferrari and Canny~\cite{ferrari1992planning}. However, it has been noted by different authors that
 analytical metrics do not guaranteee a stable grasp when executed in a real robot. 
This can be explained by the fact that these classical metrics consider assumptions 
that don't always hold true in real environments (i.e. static object-hand interaction, Coulomb friction and point contact). On the other hand, studies that consider human heuristics to guide 
grasp search have shown remarkable results, outperforming classical approaches. In
\cite{balasubramanian2014physical}, Balasubramanian observed that when humans kinestetically teach a robot how to grasp objects, they strongly tend to align the robotic hand along 
one of the object's principal axis, which later results in more robust grasps. The author
termed \textit{skewness} to the metric measuring the axis deviation. In \cite{przybylski2011human}, Przybylski et al. combine the latter metric with $\epsilon$ and use it to rank
grasps produced with GraspIt!. Berenson et al.\cite{berenson2007grasp} proposed a score combining
3 measures: $\epsilon$, object clearance and the robot relative position to the object.

%In this work we are interested in manipulation of unknown objects. Multiple approaches
%of this kind have flourished during the last few years, 
%particularly due to the advent of affordable RGB-D sensors. Since the 3D information is
%partial and noisy, classical approaches to grasp generation cannot be directly used. Rather,
%most of the current work uses heuristics to guide grasp generation based on local representation
%of the object geometry features (or global features if the object shape is approximated). In
%\cite{hsiao2010contact}, Hsiao et al. use the bounding box of the object segmented pointcloud
%to calculate grasp approach directions using a set of heuristics. We should notice that for
%most of these approaches, their effectiveness can only be verified empirically.

All the work aforementioned focus on grasp-centric metrics, whereas we stress the importance of choosing a grasp also taking into account the arm kinematics
as to encourage grasps that are easily reachable. In general, the problem of grasp planning is considered isolated from arm planning, although there are a few exceptions: Vahrenkamp et al. proposed Grasp-RRT \cite{vahrenkamp2012simultaneous} in order to perform
both grasp and arm planning combined. In a similar vein, Roa et al. also proposed an approach
that solve both problems simultaneously \cite{roa2014integrated}. Both approaches focus on \textit{reaching tasks}. Along the same lines, Berenson et al. proposed the use of Task Space Regions \cite{berenson2011task} that allow planning arm movements while also searching grasps. However, the main disadvantages of
this approach are that the object needs to be known beforehand, the task regions must be
explicitly defined by the user and it does not have a specific way to deal with multi-step tasks (planning occurs one step at a time, with no way to make the goal influence the grasp selection on an earlier step). Finally, a myriad of work exists that analyzes complex sequential manipulation tasks from a learning point of view, mostly in the form of imitation learning. However, most of these approaches focus solely on the arm motions and are naturally dependent on the
number and variety of human demonstrations available \cite{yamaguchi2015pouring,kroemer2015towards}. For tasks as simple as the ones considered in
this paper (pick-and-place and pouring), we expect that successful results can be obtained by plainly selecting the grasp to try first with a sensible metric such as $m_{ag}$.

Lastly, during recent years, deep learning approaches have flourished in diverse areas of robotics, manipulation being not the exception, with applications showcasing robots capable of picking up objects from a bin~\cite{levine2016learning}, opening doors~\cite{gu2016deep} and learning to push objects inside a crate~\cite{finn2016deep}. Our work, however, has as a main goal to provide a simple, online grasp selection strategy that do not require any kind of offline training and that can handle novel, simple objects for which several candidate grasps are generated on the fly.

%%%%%%%%%%%%%%%%%%%
%% Quick Recap
%%%%%%%%%%%%%%%%%%%
\section{Arm-and-hand metric for grasp selection}
\label{sec:RecapMetric}
In this section we briefly recap the arm-and-hand metric we will use through the rest of
this paper ($m_{ag}$), first describing each of its two component parts ($m_{a}$ and $m_{g}$) and then
their combination $m_{ag}$. A detailed description of $m_{ag}$ and the results of applying it to
simple pick-up tasks can be found in \cite{huaman2016metric}.

\subsection{Arm Metric ($m_{a}$)}
When humans perform simple reaching actions, they select a grasp
 such that their arm is comfortable at the end of the reaching movement. 
This inherently simple phenomenon, known as the \textit{end comfort effect}, has been
observed in adult humans as well as in other primates~\cite{rosenbaum1996cognition}. 

Our proposed arm-centric metric intends to capture the comfort factor
for a given grasp. Formally, for a given grasp \grasp{} applied on an object located at \Tf{w}{o}
we define our arm metric as the number of collision-free inverse kinematic solutions
that allow the robot hand to execute \grasp{}. 
\begin{equation*}
  m_{a}(\mathbf{g}) = |\mathcal{Q}| \text{ such that } \forall \mathbf{q_{i}} \in \mathcal{Q}
\begin{cases}
 \mathbf{q_{i}}\quad\text{is collision-free}\\
FK(\mathbf{q_{i}}).(\mathbf{g}.{}^{h}T_{o}) = {}^{w}T_{o} \\
\end{cases}
\end{equation*}

%^^^^^^^^^^^^^^^^^^^^^^^^^^^^^^^^^^^^^^^^^^^^^^^^^
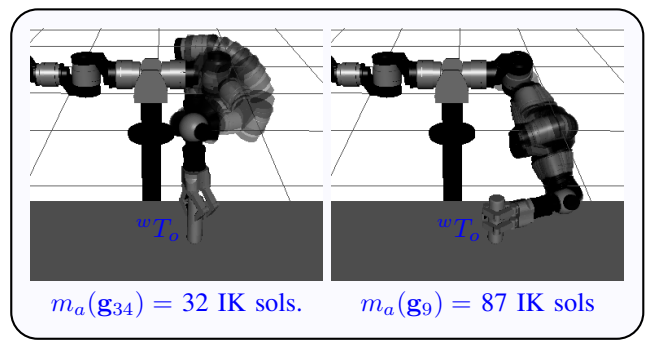
\begin{figure}[h]
 \vspace{-1.0em}
\centering
  \begin{tikzpicture}[framed,background rectangle/.style={thick,draw=black,fill=bgcolor,rounded corners=1em}]
    \tikzstyle{block} = [node distance = 3.5cm,rounded corners=.00cm,
    inner sep=.1cm, fill=bgcolor, minimum height=2em,minimum width=7em]
    %\node[block,name=cover,xshift=0cm] {\includegraphics[width=0.3\linewidth]{./figures/back_grasp_composed_2_cut.png} };
    \node[block,name=cover,xshift=0cm] {\includegraphics[width=0.45\linewidth]{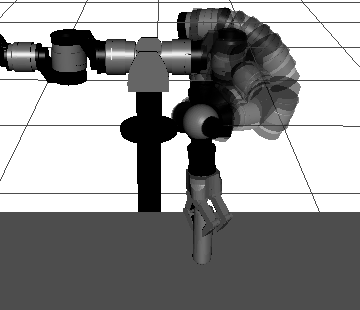} };
    \node[block,name=cover,xshift=4cm] {\includegraphics[width=0.45\linewidth]{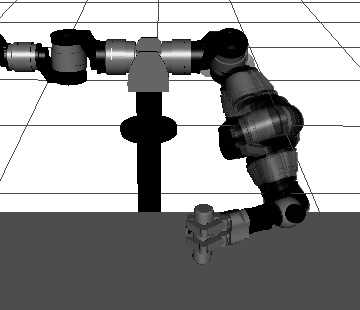} };
    \node[text=blue,xshift=-0.25cm,yshift=-1cm,font=\normalsize]{\Tf{w}{o}};
    \node[text=blue,xshift=3.75cm,yshift=-1cm,font=\normalsize]{\Tf{w}{o}};
    \node[text=blue,xshift=0cm,yshift=-2cm,font=\normalsize]{$m_{a}(\mathbf{g}_{34})=$ 32 IK sols.};
    \node[text=blue,xshift=4cm,yshift=-2cm,font=\normalsize]{$m_{a}(\mathbf{g}_{9})=$ 87 IK sols};
    %\node[text=blue,xshift=6cm,yshift=-2cm,font=\tiny]{$m_{a}$};
  \end{tikzpicture}
\vspace{-0.25em}
\caption{Visualization of $m_{a}$ calculation: In this example, there exist 80 candidate grasps that allow the
robot to grasp the object. The figure shows two grasps ($\mathbf{g}_{34}$ and $\mathbf{g}_{9}$), and their corresponding $m_{a}$. Intuitively, the grasp that comes from the side ($\mathbf{g}_{9}$) has a higher $m_{a}$ as the arm has many redundant ways to grasping the object using this grasp.}
\label{fig:armMetricViz}
\vspace{-0.5em}
\end{figure}
%^^^^^^^^^^^^^^^^^^^^^^^^^^^^^^^^^^^^^^^^^^^^^^^^^

For our specific setup, the redundant robot arm presents a standard S-R-S configuration for which
a pseudo-analytic solution is available \cite{shimizu2008analytical} given as input
an end-effector's goal pose and a free parameter $\phi \in [-\pi,\pi]$ which determines the elbow
pose. In the equation above, the initial set of inverse kinematic solutions are calculated by 
discretizing $\phi$ and evaluating which of them are collision-free.

%%%%%%%%%%%%%%%%%%%%%%%%%%%%%%%%%%%%%%%%%
\subsection{Grasp Metric ($m_{g}$)}

\begin{figure}[h]
\vspace{-1.0ex}
\centering
  \begin{tikzpicture}[framed,background rectangle/.style={thick,draw=black,fill=bgcolor,rounded corners=1em}]
  	\tikzstyle{label} = [node distance = 1.7cm,text=blue]
    \tikzstyle{mycoord} = [node distance = 0cm]
    \tikzstyle{block} = [node distance = 3.5cm,rounded corners=.00cm,
    inner sep=.1cm, fill=bgcolor, minimum height=2em,minimum width=7em]
    \node[block,name=cover] {\includegraphics[width=0.8\linewidth]{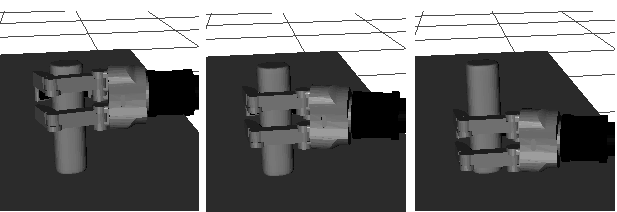} };
    \draw[->,thick,-latex,color=red!90!black,draw,line width=1mm](-2pt,-5pt)--(-32pt,-5pt);
    \draw[->,thick,-latex,color=red!90!black,draw,line width=1mm](-70pt,2pt)--(-100pt,2pt);
    \draw[->,thick,-latex,color=red!90!black,draw,line width=1mm](70pt,-11 pt)--(40pt,-11pt);
  \end{tikzpicture}
\caption{Examples of similar grasps with a different distance from the hand approach direction (red arrows) and the object center of mass}
\label{fig:graspMetric1}
\vspace{-1.0em}
\end{figure}
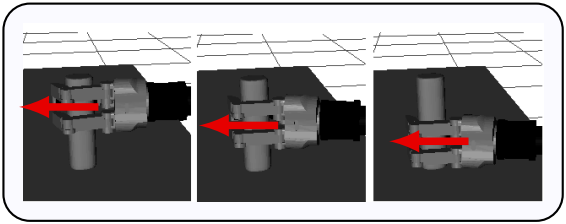

The arm-centric metric presented above only considers the arm comfort. Consider the
scenario in \fref{fig:graspMetric1}, where 3 candidate grasps are depicted for a cylindrical object
(a grasp here being parameterized by two elements: (1) The relative rigid transform of the end-effector frame with respect
to the object frame, and (2) The finger's initial joint configurations).
Let us assume that these grasps have similar $m_{a}$ values, hence they
are all deemed equally desirable. From human experience, we can all agree that the second
grasp is the most likely to be stable since the hand is closer to the center of mass of the object
being held. We incorporate this heuristic on the proposed grasp metric.

Our second metric  attempts to favor grasps that hold the object near its center of gravity. 
We propose to quantify this heuristic as the distance between the object's center of mass
and the hand's approach direction vector. We select this metric because it is easy
to calculate, as it is just the distance between a line and a point. This metric is similar
to the existing metric $B_{1}$\cite{leon2014characterization}, which measures the distance between the center of the contact polygon  and the center of mass of the object. We prefer our metric over $B_{1}$ mainly 
because our system does not provide finger contact information.

%^^^^^^^^^^^^^^^^^^^^^^^^^^^^^^^^^^^^^^^^^^^^^^^^^
%^^^^^^^^^^^^^^^^^^^^^^^^^^^^^^^^^^^^^^^^^^^^^^^^^

\subsection{Arm-Grasp Metric ($m_{ag}$)}

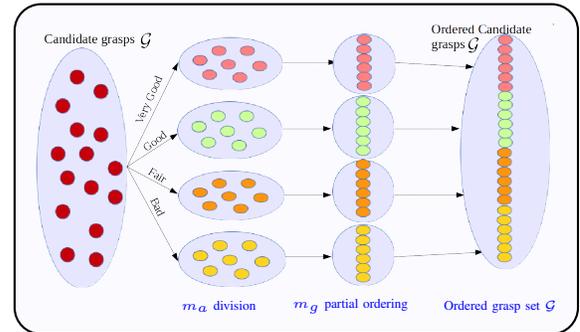
\begin{figure}[h]
\vspace{-1.0ex}
\centering
  \begin{tikzpicture}[framed,background rectangle/.style={thick,draw=black,fill=bgcolor,rounded corners=1em}]
    \tikzstyle{block} = [node distance = 3.5cm,rounded corners=.00cm,
    inner sep=.1cm, fill=bgcolor, minimum height=2em,minimum width=7em]
    \node[block,name=cover] {\includegraphics[width=0.8\linewidth]{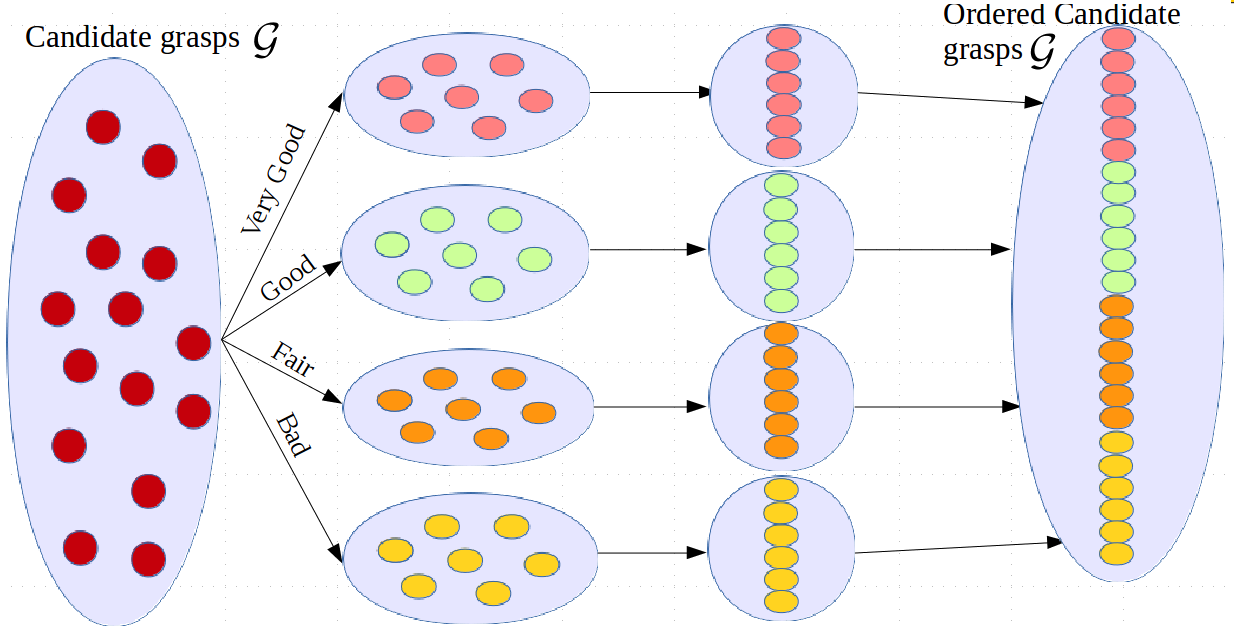} };

    \node[text=blue,xshift=-1.0cm,yshift=-2cm,font=\tiny]{$m_{a}$ division};
    \node[text=blue,xshift=0.75cm,yshift=-2cm,font=\tiny]{$m_{g}$ partial ordering};
    \node[text=blue,xshift=2.75cm,yshift=-2cm,font=\tiny]{Ordered grasp set \GraspSet{}};
  \end{tikzpicture}
\vspace{-0.5em}
\caption{Visual representation of final metric calculation}
\label{fig:caterpillar}
\vspace{-1.0em}
\end{figure}

Now that we have both metrics, we must combine them. A direct way to do this
could be using a weighted sum of both. However, both metrics have different units ($m_{a}$
 is adimensional and $m_{g}$ has length units), hence adding them 
is not straightforward. Instead, we propose to calculate $m_{ag}$ using 2 consecutive steps (illustrated in \fref{fig:caterpillar}), each
of which uses one of the metrics at a time: In the first step,
$m_{a}$ is used to divide the grasp set \GraspSet{} in 4 groups according to their $m_{a}$ quality,  whereas in the second 
step, $m_{g}$ is used to further order the grasps within each subgroup. This can be explained in simple terms as:

\begin{enumerate_tight}
\item{Calculate the mean $\mu_{a}$ and the standard deviation $\sigma_{a}$ of the arm metric ($m_{a}$) 
over all the grasps in \GraspSet{}.}
\item{Divide the grasps in 4 groups, similarly as \cite{leon2014characterization}: 
\begin{enumerate_tight}
\item{Very good $m_{a}$ quality:  $m_{a}(\mathbf{g}_{i}) > \mu_{a} + \sigma_{a}$}
\item{Good $m_{a}$ quality: $\mu< m_{a}(\mathbf{g}_{i}) <\mu_{a} + \sigma_{a}$}
\item{Fair $m_{a}$ quality: $\mu_{a}-\sigma_{a} < m_{a}(\mathbf{g}_{i}) < \mu_{a}$}
\item{Bad $m_{a}$ quality: $m_{a}(\mathbf{g}_{i}) <\mu_{a} - \sigma_{a}$}
\end{enumerate_tight}}
\item{Within each of the 4 groups, reorder the grasps according to their grasp metric $m_{g}$.}
\item{The final ordered set of grasps will contain 4 $m_{a}$-based ordered sets (very good, good, fair and bad), inside each of which grasps are ordered according to $m_{g}$}.
\end{enumerate_tight}

%^^^^^^^^^^^^^^^^^^^^^^^^^^^^^^^^^^^^^^^^^^^^^^^^^
%^^^^^^^^^^^^^^^^^^^^^^^^^^^^^^^^^^^^^^^^^^^^^^^^^

It is worth noticing that, rather than producing a numeric value, $m_{ag}$ in fact outputs
an ordering of the grasps in \GraspSet{}. In \cite{huaman2016metric} we showed that
by selecting the  grasp $g_{i}$ ranked as the best in \GraspSet{}, the average
arm motion length and end effector displacement entailed was shorter than when selecting
other candidate grasp. In the following sections we will analyze what is the best way to
use this metric for 2-step tasks such as Pick-and-Place (\sref{sec:PickAndPlaceTasks}) and Pouring (Section \sref{sec:PouringTasks}).

%%%%%%%%%%%%%%%%%%%%%
%% Task Definitions
%%%%%%%%%%%%%%%%%%%%%
%\section{Tasks Definitions}
%\label{sec:TaskDefinitions}
%\input{taskDefinitions}%%%%%%%

%%%%%%%%%%%%%%%%%%%%%%%%%%%%
%% Pick-and-Place tasks
%%%%%%%%%%%%%%%%%%%%%%%%%%%%
\section{Pick-and-Place Tasks}
\label{sec:PickAndPlaceTasks}

\subsection{Task Definition} 
Given an object~\Object~at a start pose~\Tf{w}{s}~, the robot must
reposition~\Object~to a 3D goal position~\Pos{w}{g}~keeping the object upright. Both start and goal states must be
inside the reachable workspace of the robot arm being used. The task is considered successful if 
at the end of the pick-and-place execution the object~\Object~is set at~\Pos{w}{g}~with a small margin of position error.

\begin{figure}[h]
\vspace{-2ex}
\centering
\begin{tikzpicture}[framed,background rectangle/.style={thick,draw=black,fill=bgcolor,rounded corners=1em}]
\node[]{\includegraphics[width=0.45\textwidth]{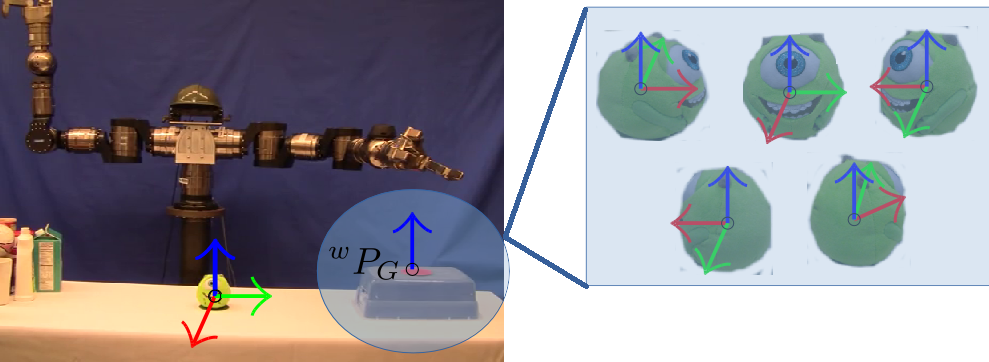}};
\node[text=black,xshift=-2.75cm,yshift=-1.0cm,font=\normalsize]{\Tf{w}{s}};
\end{tikzpicture}
\caption{Pick-and-place task with not-fully constrained goal state: Many possible yaw orientations could be selected.}
\label{fig:yawLiberty}
\vspace{-1.0em}
\end{figure}

Notice that the goal state is not fully constrained (\fref{fig:yawLiberty}). Specifically,
the yaw orientation of~\Object~can adopt many possible values. This presents a challenge for testing our metric $m_{ag}$, since for 
it to be evaluated we need a full 6D goal pose (\Tf{w}{g}) to be defined. We chose to use this task description over a fully-constrained one as this is a very common pick-and-place variation that routinely appears in household scenarios..

In order to measure $m_{ag}$ at the goal state, we propose to use what
we term a \textit{goal pose guess} (6D) for each 
candidate grasp, such that this can be used as an estimation of the likely 6D pose of~\Object~ at the goal sate. In the next section we explain how
to generate these \textit{goal pose guesses} based on a simple human heuristic.

\subsection{Generating likely goal pose guesses  (${}^{w}P_{g}\rightarrow$\Tf{w}{g})}
\label{subsec:Pap_generateGuessGoalPose}
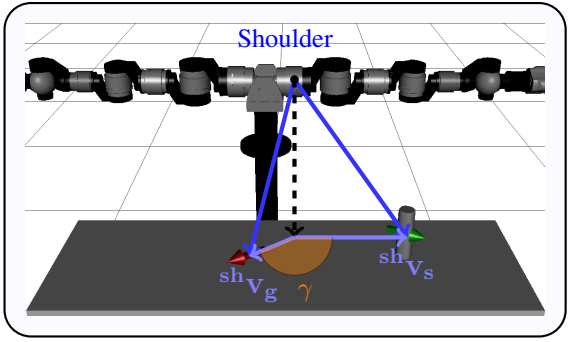
\begin{figure}
\vspace{-2ex}
\centering
\begin{tikzpicture}[framed,background rectangle/.style={thick,draw=black,fill=bgcolor,rounded corners=1em}]
\node[]{ \includegraphics[width=0.8\linewidth]{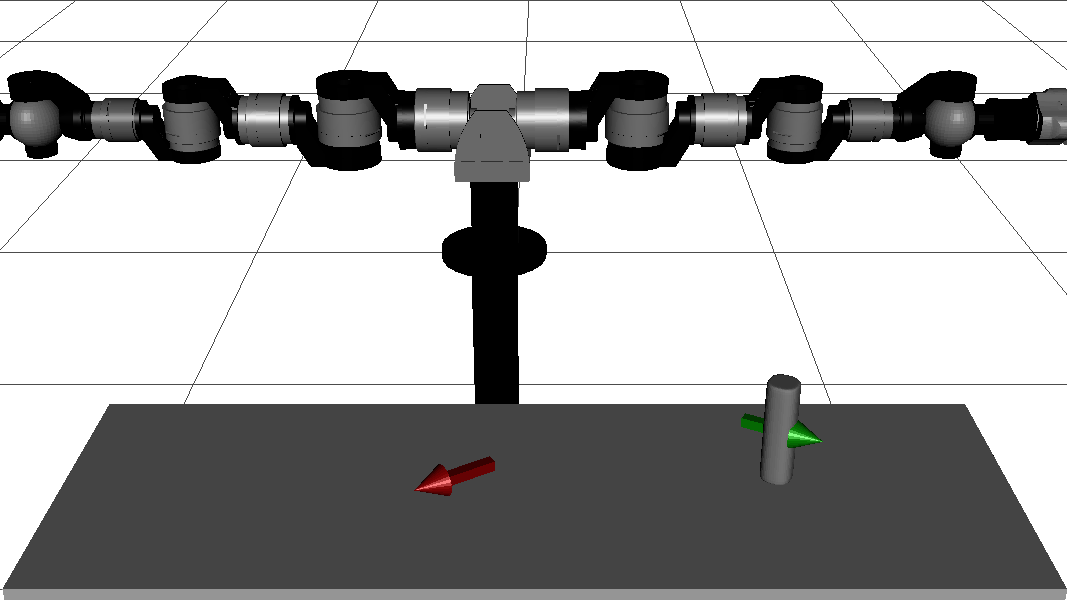} };
\coordinate (Shoulder) at (0.125, 1.2 ,0);
\coordinate (ShoulderT) at ($(Shoulder) + (0.0, -2.1)$);
\coordinate (Start) at ($(Shoulder) + (1.5, -2.1)$);
\coordinate (Goal) at ($(Shoulder) + (-0.6, -2.35)$);
\tdplotsetrotatedcoordsorigin{(Shoulder)}
\node[text=blue,xshift=0cm,yshift=1.75cm]{Shoulder};
\draw[ultra thick,tdplot_rotated_coords,color=blue!80!white,->](Shoulder) -- (Goal) node [anchor= south]{};
\draw[ultra thick,tdplot_rotated_coords,color=blue!80!white,->](Shoulder) -- (Start) node [anchor= south]{};
\draw[ultra thick,dashed,->,tdplot_rotated_coords,color=black](Shoulder) -- (ShoulderT) node [anchor= north]{};
\draw[ultra thick,tdplot_rotated_coords,color=blue!50!white,->](ShoulderT) -- (Goal) node [anchor= north]{\large $\mathbf{{}^{sh}v_{g}}$};
\draw[ultra thick,tdplot_rotated_coords,color=blue!50!white,->](ShoulderT) -- (Start) node [anchor= north]{\large $\mathbf{{}^{sh}v_{s}}$};
\tkzMarkAngle[fill=orange,size=0.5cm,opacity=0.4](Goal,ShoulderT,Start)
\tkzLabelAngle[pos=0.75,color=orange](Goal,ShoulderT,Start){$\gamma$}
\draw[fill,black] (Shoulder) circle (0.05cm);
\end{tikzpicture}
\caption{Visualization of calculation of yaw angle limit $\gamma$ (green marker is the start state, red marks the goal position).}
\label{fig:papRelaxed_YaxMax}
\vspace{-2ex}
\end{figure}

We designed a simple heuristic to generate \textit{goal pose guesses} for candidate grasps in pick-and-place tasks. We assume that we have a set of 
candidate grasps \GraspSet{} feasible to execute on~\Object~at \Tf{w}{s}, but yet untested in the goal
position $^{w}P_{g}$ as it is not fully defined. For each grasp $\mathbf{g}_{i} \in$ \GraspSet{}, we define their corresponding
\textit{goal pose guess} as follows:

\begin{itemize_tight}
\item{Calculate a referential rotation ($\gamma$) from the start pose \Tf{w}{s} to the goal position $^{w}P_{g}$. We do this
by generating vectors originating in the shoulder point and pointing to both the 3D origin of \Tf{w}{s} and to $^{w}P_{g}$. The referential rotation $\gamma$ is the angle between these two vectors projected
on the table plane (considering z as the up direction this is in fact a yaw angle). \fref{fig:papRelaxed_YaxMax} shows a visualization of $\gamma$ for a sample problem.}
\item{We use the referential rotation calculated above as a maximum limit for the goal's relative
rotation with respect to \Tf{w}{s}. In general, we assume that for manipulation tasks,
only the minimum effort necessary will be used (also known as the human "economical principle'' \cite{nagasaki1989asymmetric}). Under this assumption, a pick-and-place operation will apply a rotation on the object only when it
is necessary.}
\item{For each candidate grasp \grasp{i}, we set~\Object~ in a goal pose such that the relative rotation varies
between 0 (minimum rotation) and $\gamma$. We discretize this interval in a small number of 
samples and test if \grasp{i} is feasible at each of them. If so, we stop the testing and store
this \textit{goal pose guess} for future use in the grasp selection process. Notice that we
start searching for pose guesses starting from zero, as we assume that rotations are minimum.
Once a feasible pose is found, the search stops.}
\end{itemize_tight}

The algorithmic version of the heuristic is depicted in Algorithm \ref{alg:papr_getGoalPoseGuess}.
As it is shown, the output of this procedure is a \textit{guess goal pose} per each candidate grasp. We use these guess poses instead of the original goal positions to calculate our metric $m_{ag}$ and evaluate it accordingly. 

\begin{algorithm}
\DontPrintSemicolon
\KwIn{\Tf{w}{s}, ${}^{w}P_{g}$, \Object, \GraspSet{}}
\KwOut{$\mathcal{T}_{G}$: Set of goal pose guesses for each \grasp{i} $\in$ \GraspSet{} }
\SetKwFunction{generateGrasps}{generate\_Grasps}
\SetKwFunction{setPose}{set\_Pose}
\SetKwFunction{erase}{.erase}
\SetKwFunction{existIKSol}{exist\_IK\_sol}
\SetKwFunction{projectToTable}{project\_to\_table}
\SetKwFunction{Rotz}{Rot$_{z}$}
\SetKwFunction{FALSE}{false}
\SetKwFunction{pushback}{.push\_back}
\SetKwFunction{TRUE}{true}
\BlankLine

\tcc{Calculate referential max. orientation $\gamma$ for the goal pose}
$^{sh}v_{s}$ $\leftarrow$ \Tf{w}{s}.trans() - $^{w}P_{\text{sh}}$\quad \tcc{$^{w}P_{\text{sh}}$ is the shoulder position}
${}^{sh}v_{g}$ $\leftarrow$ ${}^{w}P_{g} - {}^{w}P_{\text{sh}}$ \;
{${}^{sh}v_{s}$} $\leftarrow$ \projectToTable{${}^{sh}v_{s}$} \;
{${}^{sh}v_{g}$} $\leftarrow$ \projectToTable{${}^{sh}v_{g}$} \;
$\gamma$ $\leftarrow$ \Rotz{${}^{sh}v_{s}$, ${}^{sh}v_{g}$}\;

\tcc{Assign a possible goal pose to each grasp \grasp{i} }
\ForEach{ \grasp{i} $\in$ \GraspSet{} } {
  \Tf{w}{g} $\leftarrow$ NULL \;
  \ForEach{ i $\in$ $[0..N]$}{
    $\gamma_{i} \leftarrow  (\gamma/N)\cdot i$ \;
    \Tf{w}{g}$^{*}$ $\leftarrow$ \Rotz{$\gamma_{i}$}$\cdot$ \Tf{w}{s}.rot() \;
    \tcc{Put object in this pose and evaluate if grasp reaches}
    \setPose{\Object,\Tf{w}{g}}\;
    \If{\existIKSol{\grasp{i},\Tf{w}{g}} {\bf is} \TRUE }{
      \Tf{w}{g} $\leftarrow$ \Tf{w}{g}$^{*}$\;
      break \;
    }
  }
    $\mathcal{T}_{G}$\pushback{\Tf{w}{g} } \; 
}
  \Return $\mathcal{T}_{G}$ \;
\caption{get\_Goal\_Pose\_Guesses}
\label{alg:papr_getGoalPoseGuess}
\end{algorithm}

%***************************
%%%

%%%%%%%%%%%%%%%%%%%%%%%%%%%%%%%%%%%%%%%%%%%%%%%%%%%%%%%%
\subsection{Calculating $m_{ag}$ as an average for both start and goal states}
Once we calculated our goal guess poses, we can measure $m_{ag}$ at both
the start and goal state. However, we also would like to get a measurement
that considers both of them at the same time. As we pointed out, our metric
produces as an output an ordering rather than a numerical value, so a
standard averaging is not possible. We observe, however, that although our
metric is composed of two components, $m_{a}$ and $m_{g}$, only $m_{a}$ changes
when evaluated at different object's poses ($m_{g}$ remains the same as
the grasps are rigid). Given this, a simple way to calculate an average
version of $m_{ag}$ consists on averaging the $m_{a}$ values at the start and
goal states and then use this average in combination with the non-changing $m_{g}$ to produce the final ordering, incorporating then both the start and
goal information in the process. The algorithmic version corresponding to this
explanation can be seen in \aref{alg:avgMetricPaP}.

\begin{algorithm}
\DontPrintSemicolon
\KwIn{\Tf{w}{s}, \Tf{w}{g}, \Object, \GraspSet{} }
\KwOut{Prioritized set of grasps \GraspSet{}$^{*}$ }
\SetKwFunction{pushback}{.push\_back}
\SetKwFunction{featureScale}{feature\_scale}
\SetKwFunction{getMaMgMetric}{get\_$m_{a}$\_$m_{g}$\_metric}
\SetKwFunction{getMagMetric}{get\_$m_{ag}$\_metric}
%\BlankLine
\tcc{Generate the arm metric at start and goal pose}
 $($\ArmMetricSet$^{S}$, \GraspMetricSet $)$  $\leftarrow$ \getMaMgMetric{\Tf{w}{s}, \Object, \GraspSet{} }\;
 $($\ArmMetricSet$^{G}$, \GraspMetricSet $)$ $\leftarrow$ \getMaMgMetric{\Tf{w}{g}, \Object, \GraspSet{} }\;

\tcc{Normalize the arm metrics values $\in [0,1]$}
 \ArmMetricSet$^{S}$ $\leftarrow$ \featureScale{\ArmMetricSet$^{S}$}\;
 \ArmMetricSet$^{G}$ $\leftarrow$ \featureScale{\ArmMetricSet$^{G}$}\;
\tcc{\ArmMetricSet$^{SG}$: Added arm metric using both start and goal arm metric values}
\ForEach{ \grasp{i} $\in$ \GraspSet{} }{
 \ArmMetricSet$^{SG}$\pushback{ \ArmMetricSet$^{S}[i]$ +  \ArmMetricSet$^{G}[i]$ } \;
}
\tcc{Order the grasps the way we used to do, using the combined arm metric}
\Return \getMagMetric{ \ArmMetricSet$^{SG}$, \GraspMetricSet, \GraspSet{}}
\caption{get\_Average\_$m_{ag}$\_metric}
\label{alg:avgMetricPaP}
\end{algorithm}

%%%%%%%%%%%%%%%%%%%%%%%%%%%%%%%%%%%%%%%%%%%%%%%%%%%%%%%%%%%%%
\subsection{Evaluation of $m_{ag}$ at start, goal and as an average}
\label{subsec:EvalPaP}
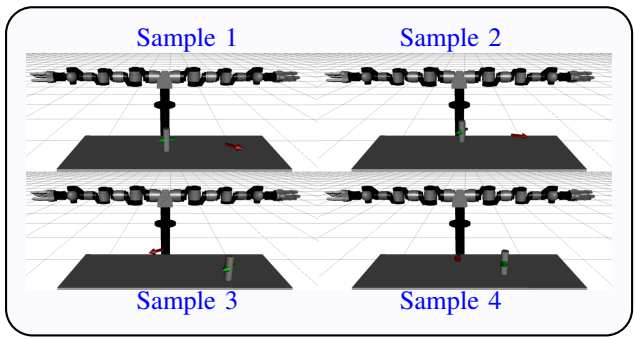
\begin{figure}
\centering
\begin{tikzpicture}[framed,background rectangle/.style={thick,draw=black,fill=bgcolor,rounded corners=1em}]
\node[]{ \includegraphics[width=0.9\linewidth]{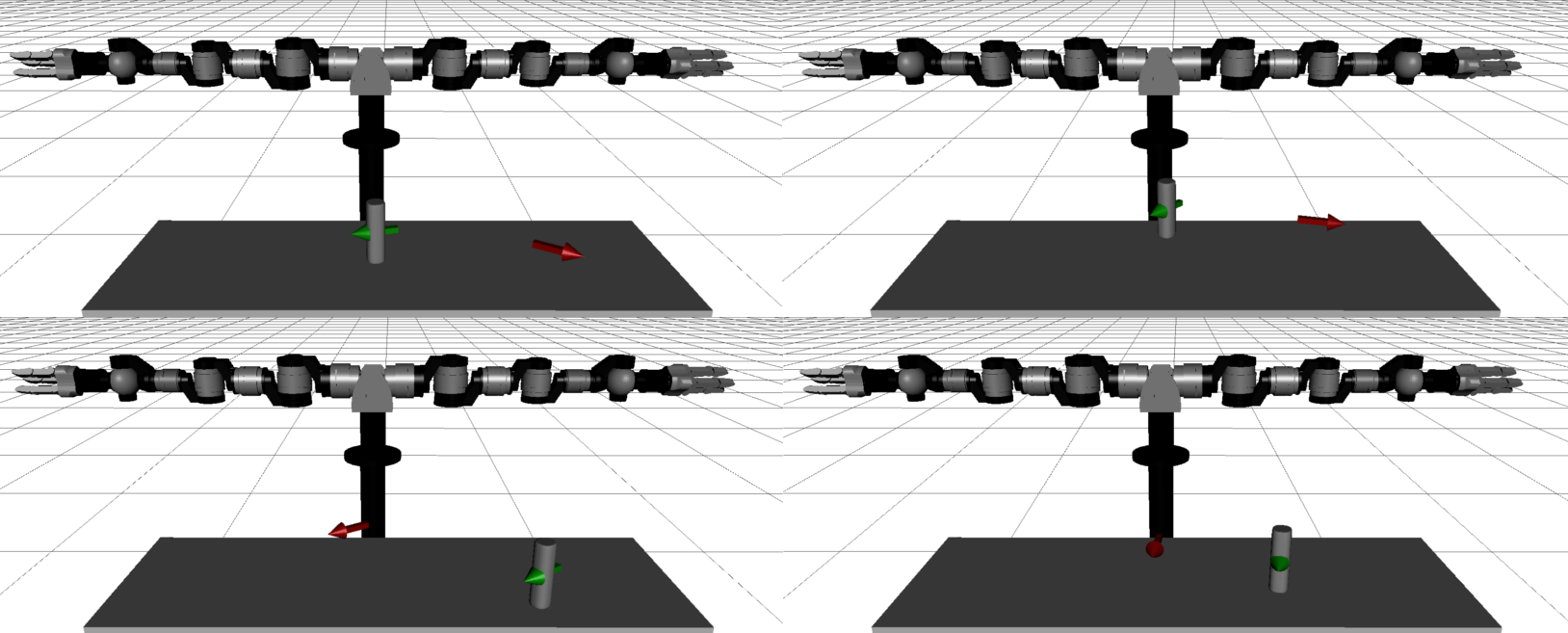} };
\node[text=blue,xshift=-1.75cm,yshift=1.75cm]{Sample 1};
\node[text=blue,xshift=1.75cm,yshift=1.75cm]{Sample 2};
\node[text=blue,xshift=-1.75cm,yshift=-1.75cm]{Sample 3};
\node[text=blue,xshift=1.75cm,yshift=-1.75cm]{Sample 4};
\end{tikzpicture}
\caption{Samples random cases to evaluate the proposed grasp prioritization metrics in tabletop simulation (green and red markers indicate random start pose and goal position respectively).}
\label{fig:pap_simSample}
%\vspace{-2.0em}
\end{figure}

\begin{figure}
\centering
\begin{tikzpicture}[framed,background rectangle/.style={thick,draw=black,fill=bgcolor,rounded corners=1em}]
\node[]{ \includegraphics[width=0.8\linewidth]{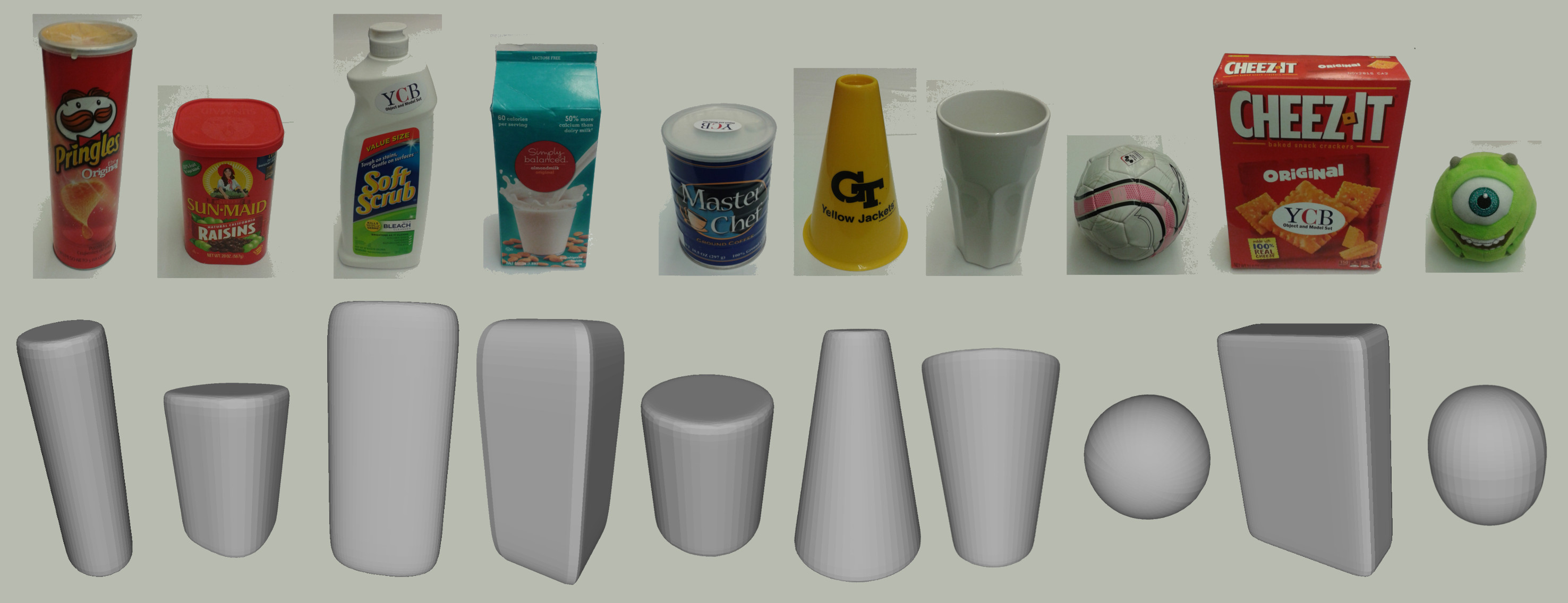} };
\end{tikzpicture}
\caption{Objects used in real experiments (Subsets of these were used in simulation via their mesh counterparts)}
\label{fig:objectsUsed}
%\vspace{-2.0em}
\end{figure}

We evaluate our metric $m_{ag}$ in a series of random pick-and-place
experiments on simulation. Using a simple tabletop scenario, such
as the one shown in \fref{fig:pap_simSample} we generate a set of candidate grasps \GraspSet{} (using the method explained in \cite{huaman2015exploiting})  and select which grasp to use according to our metric $m_{ag}$ measured
under 3 modalities: Start state, goal guess state and average.  
We performed 250 random experiments per each of 6 objects evaluated
as to make sure the results were not affected by the object's geometry
(\fref{fig:objectsUsed}).

% Table: Pringles and raisins
%%%

%%%

The results of the simulation experiments are shown in Table \ref{tab:papr_simResultsTable}. We use 3 metrics to compare the performance of the grasps selected
under $m_{ag}$ measured at different steps:

\begin{itemize_tight}
\item{\textit{Success rate:} The main point of ranking the grasps is to avoid having to try multiple grasps
before finding one that produces a feasible arm motion. The success rate measures if the
grasp ranked the best produces a solution (no further grasps are tried).}
\item{\textit{Planning time:} Ideally, the grasp selected must be easy to reach and execute, hence short
planning times should be expected.}
\item{\textit{End Effector displacement:} Related to planning time. Easy arm motions should imply short end-effector translation in the workspace.}
\end{itemize_tight}

\begin{table}[h]
\scriptsize
\setlength{\tabcolsep}{3pt}
\centering
\caption{Simulation results of randomized on-table pick-up scenarios (250) for each object}
\label{tab:papr_simResultsTable}
\begin{tabular}{l |c c c|c c c|c c c}
\topline
\headcol \textbf{Object} & \multicolumn{3}{c|}{\textbf{Success}} & \multicolumn{3}{c}{\textbf{Hand Disp.(m)}} &  \multicolumn{3}{c}{\textbf{Plan time (s)}} \\
\headcol  & \textbf{Goal} & \textbf{Start} & \textbf{Avg} & \textbf{Goal} & \textbf{Start} & \textbf{Avg} & \textbf{Goal} & \textbf{Start} & \textbf{Avg} \\
\midline
\multirow{1}{*}{Master Chef} & 96\% & 76\%   & {\color{blue}96\%} & {\color{blue}2.62} & 2.65 & 2.64 & {\color{blue}1.56} & 1.71 & 1.69 \\
% & 100/246 & 87/246 & 195/246      & 2.85 & 2.49 & 2.65 & 2.65 & 1.99 & 1.94 \\
\cmidrule{1-10} 
\multirow{1}{*}{Green plushie} & 93\% & 77\% & {\color{blue}93\%} & {\color{blue}2.50} & 2.61 & 2.55 & {\color{blue}1.47} & 1.55 & 1.52  \\
 %& 71/250 & 90/250 & 213/250 & 2.80 & 2.64 & 2.59 & 1.78 & 1.30 & 1.52 \\
\cmidrule{1-10}
\multirow{1}{*}{Pringles} & 94\% & 79\% & {\color{blue}96\%} & 2.44 & 2.46 & {\color{blue}2.42} & {\color{blue}1.55} & 2.11 & 1.73  \\
 %& 112/250 & 106/250 & 202/250 & 2.95 & 2.56 & 2.73 & 5.24 & 1.62 & 2.24 \\
\cmidrule{1-10}
\multirow{1}{*}{Soft Scrub} & 92\% & 80\% & {\color{blue}93\%} & 2.41 & 2.43 & {\color{blue}2.41} & {\color{blue}1.74} & 2.93 & 1.95 \\
%& 91/250 & 97/250 & 190/250 &  2.82 & 2.71 & 2.71 & 4.89 & 1.54 & 2.04\\
\cmidrule{1-10}
\multirow{1}{*}{Sun maid} & 88\%  & 69\% & {\color{blue}88\%} & {\color{blue}2.58} & 2.79 & 2.73 & {\color{blue}1.39} & 1.55 & 1.42 \\
%&  82/249 & 91/249 & 171/249 & 2.99 & 2.81 & 2.74 & 2.26 & 1.46 & 1.56 \\
\cmidrule{1-10}
\multirow{1}{*}{Yellow cone} & 83\% & 72\% & {\color{blue}85\%} & 2.52 & 2.54 & {\color{blue}2.51} & {\color{blue}1.79} & 2.05 & 1.88 \\
%& 103/250 & 69/250 & 192/250 & 2.89 & 2.48 & 2.63 & 3.90 & 2.03 & 2.41 \\
\bottomrule
\end{tabular}
\vspace{-0.5em}
\end{table}

The following observations are made from the aforementioned results table:

\begin{itemize_tight}
\item{The average metric presents higher success rates for the objects evaluated, with the metric measured at the goal state coming a close second. Interestingly, the metric measured at the start pose produce the lowest success rates.}
\item{The end effector distance traveled during the pick-and-place task is
in general shorter for the metric measured at the goal and average cases.}
\item{The planning times are consistently lower for the metric measured at the
goal state, although the difference is very small with respect to the other two cases.}
\end{itemize_tight}

In general, we could say that using the $m_{ag}$  metric measured at either the goal state or as an average of the start and the goal produce good results. To
give a better idea of the advantage of using this metric with respect to 
not using any metric at all, please refer to \tref{tab:papr_improv}.

\begin{table}[h]
\scriptsize
\setlength{\tabcolsep}{3pt}
\centering
\caption{Comparison of performance between grasp deemed {\color{blue}best} and {\color{red}worst} with respect to the metric $m_{ag}$ measured at the ``average'' state.}
\label{tab:papr_improv}
\begin{tabular}{l |c|c|c}
\topline
\headcol \textbf{Object} & \textbf{Success} &\textbf{Hand Disp.(m)} & \textbf{Plan time (s)} \\
\midline
\multirow{2}{*}{Master Chef} & {\color{blue}96\%} & {\color{blue}2.62} & {\color{blue}1.56}\\
 & {\color{red}41\%}  & {\color{red}2.85} & {\color{red}2.65}  \\
\cmidrule{1-4} 
\multirow{2}{*}{Green plushie} & {\color{blue}93\%}  & {\color{blue}2.50}  & {\color{blue}1.47}   \\
 & {\color{red}28\%}  & {\color{red}2.80}  & {\color{red}1.78}  \\
\cmidrule{1-4}
\multirow{2}{*}{Pringles} & {\color{blue}94\%}  & {\color{blue}2.44}  & {\color{blue}1.55}   \\
 & {\color{red}45\%}  & {\color{red}2.95} & {\color{red}5.24}  \\
\cmidrule{1-4}
\multirow{2}{*}{Soft Scrub} & {\color{blue}92\%}  & {\color{blue}2.41}  & {\color{blue}1.74} \\
& {\color{red}36\%}  &  {\color{red}2.82}  & {\color{red}4.89} \\
\cmidrule{1-4}
\multirow{2}{*}{Sun maid} & {\color{blue}88\%}  & {\color{blue}2.58} & {\color{blue}1.39} \\
&  {\color{red}33\%}  & {\color{red}2.99}  & {\color{red}2.26}  \\
\cmidrule{1-4}
\multirow{2}{*}{Yellow cone} & {\color{blue}83\%}  & {\color{blue}2.52} & {\color{blue}1.79} \\
& {\color{red}41\%} & {\color{red}2.89}  & {\color{red}3.90}  \\
\bottomrule
\end{tabular}
\vspace{0.0em}
\end{table}

\begin{figure}[h]
\centering
\begin{tikzpicture}[framed,background rectangle/.style={thick,draw=black,fill=bgcolor,rounded corners=1em}]
\node[]{\includegraphics[width=0.45\textwidth]{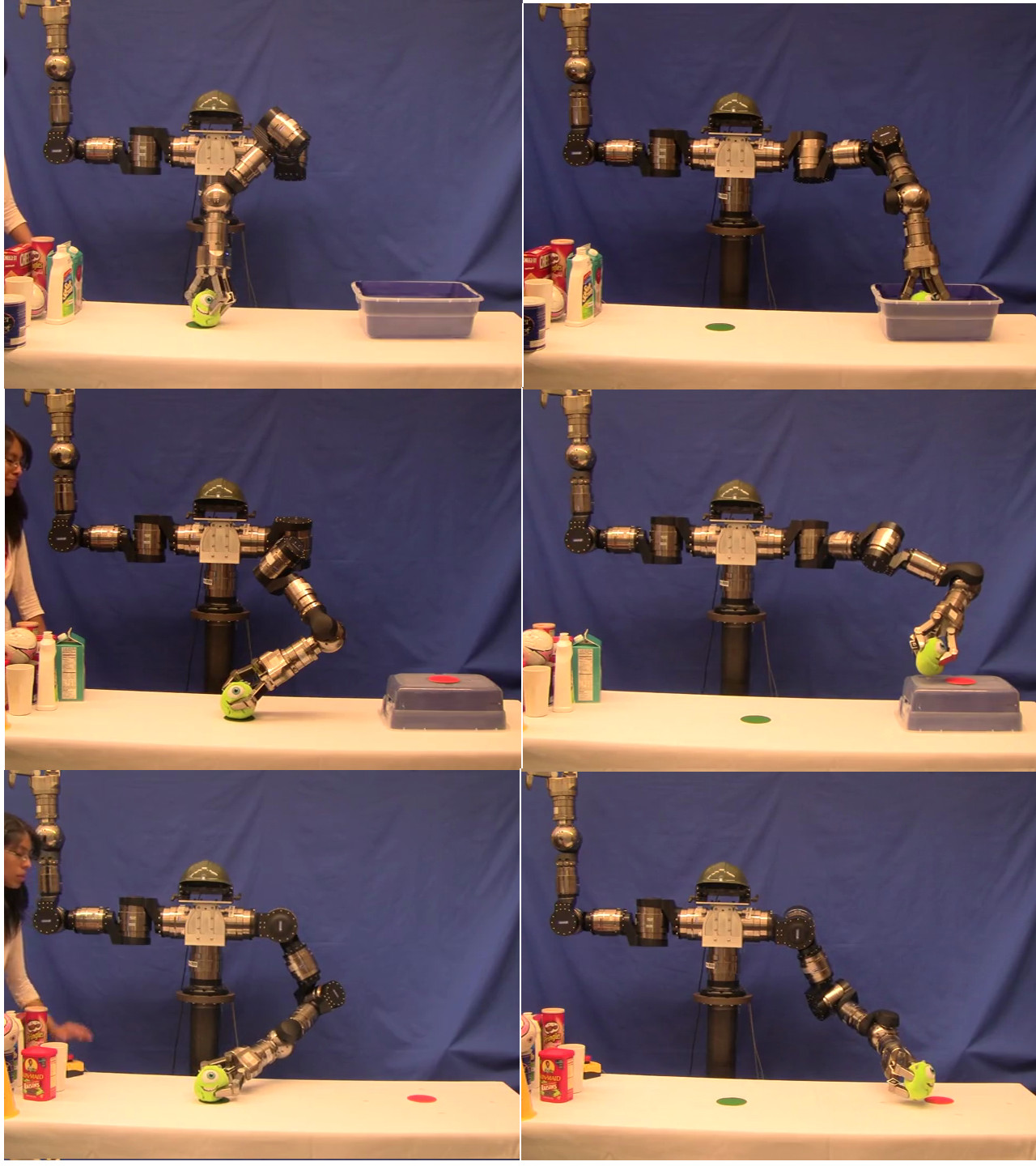}};
%\draw[line width=0.75mm, white] (-4,0) -- (4,0);
\node[text=blue,xshift=0cm,yshift=4.25cm,fill=white,fill opacity=0.5]{~~~~~~~~~~~~~Put plushie inside the box~~~~~~~~~~~~~~~};
\node[text=blue,xshift=-2.0cm,yshift=1.75cm]{Start};
\node[text=blue,xshift=2.0cm,yshift=1.75cm]{Goal};
\node[text=blue,xshift=0cm,yshift=1.25cm,fill=white,fill opacity=0.5]{~~~~~~~~~~~~~Put plushie inside the  box~~~~~~~~~~~~~~};
\node[text=blue,xshift=-2.0cm,yshift=-1.25cm]{Start};
\node[text=blue,xshift=2.0cm,yshift=-1.25cm]{Goal};
\node[text=blue,xshift=0cm,yshift=-1.75cm,fill=white,fill opacity=0.5]{~~~~~Put plushie somewhere else on the table ~~~~~~};
\node[text=blue,xshift=-2.0cm,yshift=-4.25cm]{Start};
\node[text=blue,xshift=2.0cm,yshift=-4.25cm]{Goal};
\end{tikzpicture}
\caption{Pick-and-place tasks: According to the task, the grasp selected varies}
\label{fig:pap_plushie}
\vspace{-2ex}
\end{figure}

\begin{figure}[h]
\centering
\begin{tikzpicture}[framed,background rectangle/.style={thick,draw=black,fill=bgcolor,rounded corners=1em}]
\node[]{\includegraphics[width=0.45\textwidth]{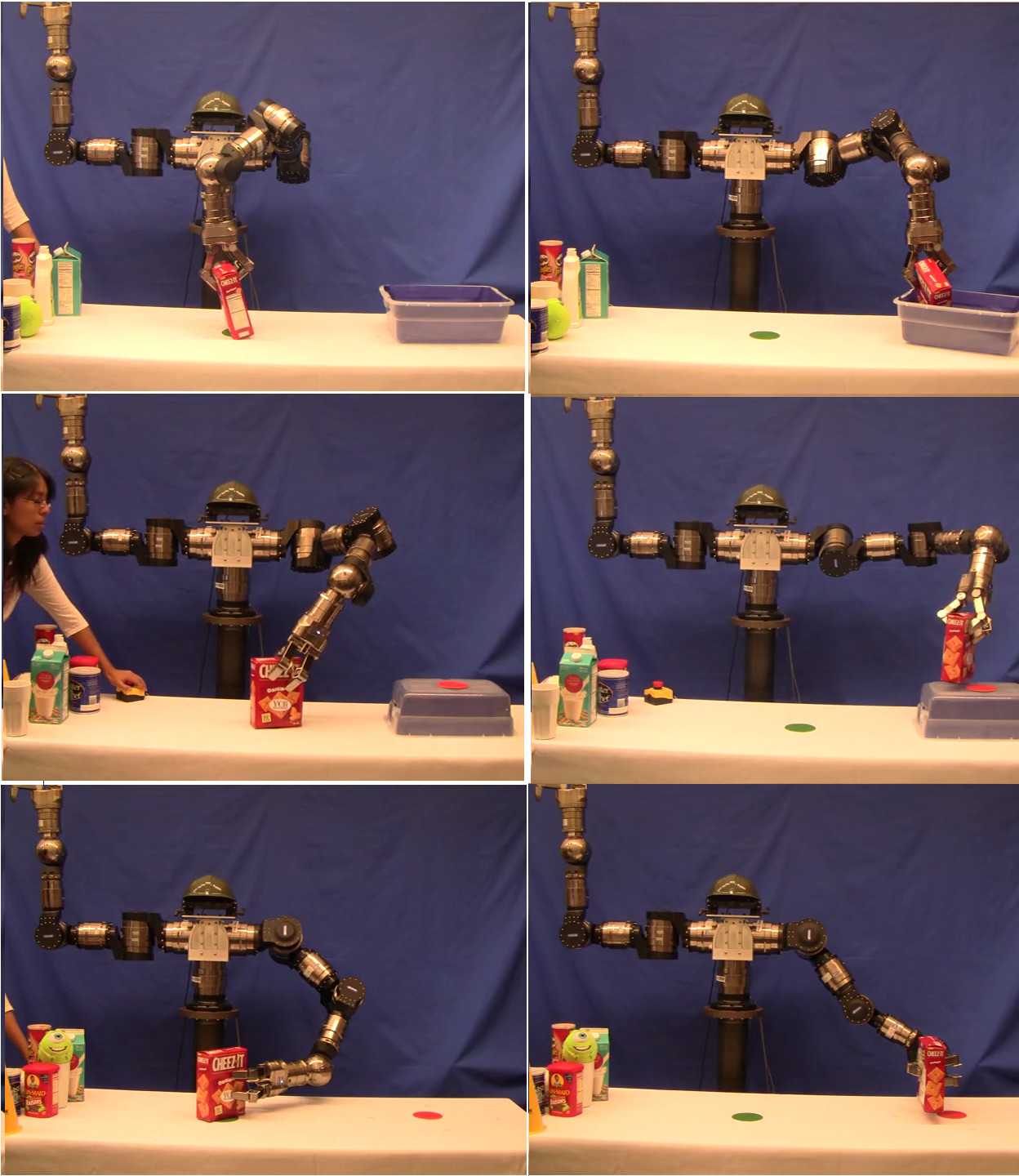}};
%\draw[line width=0.75mm, white] (-4,0) -- (4,0);
\node[text=blue,xshift=0cm,yshift=4.25cm,fill=white,fill opacity=0.5]{~~~~~~~~~~~~~Put Cheezit inside the box~~~~~~~~~~~~~~~};
\node[text=blue,xshift=-2.0cm,yshift=1.75cm]{Start};
\node[text=blue,xshift=2.0cm,yshift=1.75cm]{Goal};
\node[text=blue,xshift=0cm,yshift=1.25cm,fill=white,fill opacity=0.5]{~~~~~~~~~~~~~Put Cheezit inside the  box~~~~~~~~~~~~~~};
\node[text=blue,xshift=-2.0cm,yshift=-1.25cm]{Start};
\node[text=blue,xshift=2.0cm,yshift=-1.25cm]{Goal};
\node[text=blue,xshift=0cm,yshift=-1.75cm,fill=white,fill opacity=0.5]{~~~~~Put Cheezit somewhere else on the table ~~~~~~};
\node[text=blue,xshift=-2.0cm,yshift=-4.25cm]{Start};
\node[text=blue,xshift=2.0cm,yshift=-4.25cm]{Goal};
\end{tikzpicture}
\caption{Pick-and-place tasks: According to the task, the grasp selected varies}
\label{fig:pap_cheezit}
\vspace{-0.5em}
\end{figure}

To illustrate the aplicability of our approach, we tested it in our robotic platform, a Schunk LWA4 bimanual
manipulator. The video accompanying this paper shows the pick-and-place tasks being executed for 3 different
scenarios (place object inside a box, on top of a box or on the table surface) with 9 objects
of different geometries. \fref{fig:pap_plushie} and \fref{fig:pap_cheezit} show the start and goal
states for 2 of the evaluated objects in each of the 3 pick-and-place experiment variations. The grasps
used in each example were selected by using the average $m_{ag}$ discussed earlier. An interesting observation that can be deduced from both of these images is the fact that, even when the object geometries are different, the
grasps selected by the robot can be described as visually similar: When the object must be put inside a box, the grasp chosen is one coming from the top, likely to avoid possible collisions with the box walls. In the
on-box case, the grasp chosen has the arm coming at an angle, which makes the goal configuration more comfortable. Finally in the simplest on-table case, the grasp comes from the side, which corresponds to a relaxed
arm configuration. We consider that this is a very good advantage of using simple measurements such as $m_{ag}$: Although the arm motion planning is sampling-based (RRT), the grasp selection introduces a certain level of determinism since the grasp selected will always be the same as long as the environmental conditions are similar.

%-------------------------
% PaP plushie
%-------------------------

%-------------------------
% PaP cheezit
%-------------------------

\section{Pouring Tasks}
\label{sec:PouringTasks}
\subsection{Task Definition}
Given a pouring object~\Object{P}~in a starting pose~\Tf{w}{s}~and a second container object~\Object{R}~, the goal of the task is to position~\Object{P}~over~\Object{R}~such that transfer of contents from~\Object{P}~to~\Object{R}~is feasible. In a geometrical context, pouring in this paper is considered as a state
in which~\Object{P}~is placed above~\Object{R}~and its main principal axis form an angle $\in [0,-30]$ with the horizontal, such that~\Object{P}~is effectively pointing down. 

\begin{figure}[h]
\centering
\begin{tikzpicture}[framed,background rectangle/.style={thick,draw=black,fill=bgcolor,rounded corners=1em}]
\node[name=reach]{\includegraphics[width=0.3\linewidth]{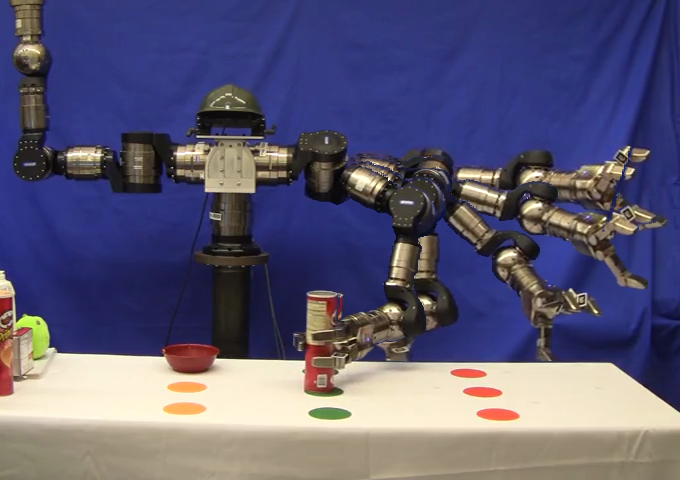}};
\node[name=transport, xshift=2.75cm]{\includegraphics[width=0.3\linewidth]{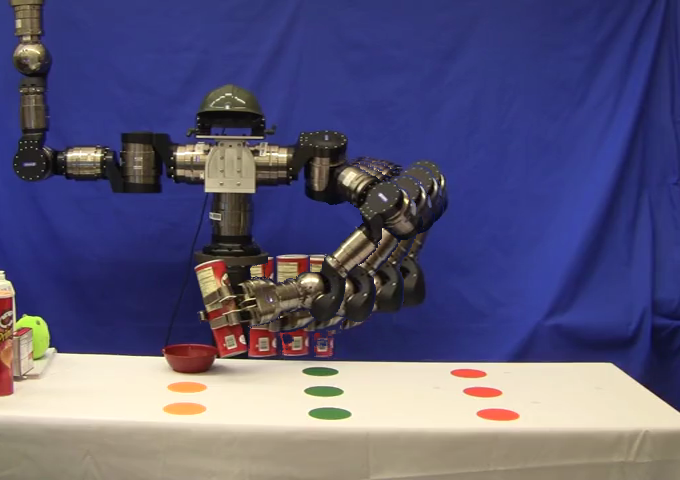}};
\node[name=tilt,xshift=5.5cm]{\includegraphics[width=0.3\linewidth]{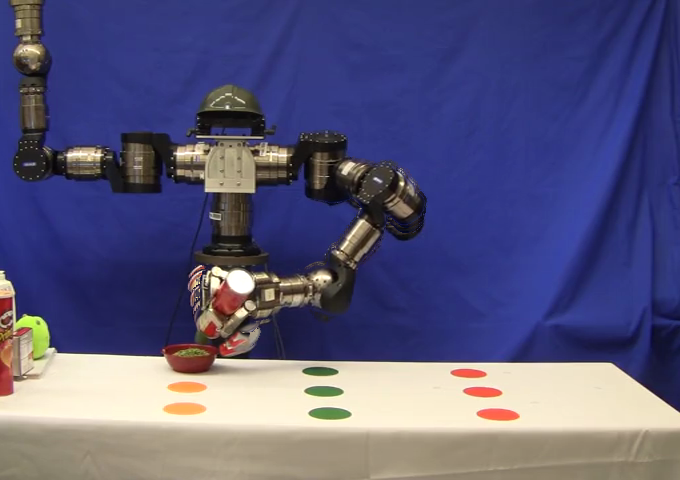}};
\node[text=blue, xshift=0cm,yshift=-1.25cm]{Reaching};
\node[text=blue, xshift=2.75cm,yshift=-1.25cm]{Transporting};
\node[text=blue, xshift=5.5cm,yshift=-1.25cm]{Tilting};
\end{tikzpicture}
\caption{Standard pouring task. In this example, the Pringles container is \Object{P} and the red bowl is the receiver \Object{R}. Multiple possible poses for
the Pringles tilting around the bowl are valid.}
\label{fig:pouring_task_view}
\vspace{-2ex}
\end{figure}
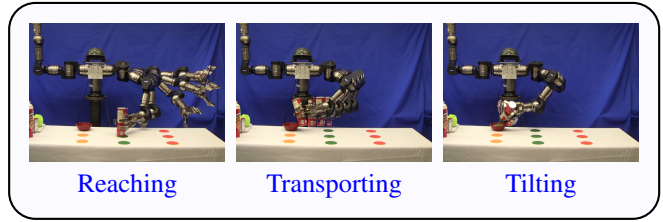

In the pouring task we found that, analogously to the pick-and-place task, the
goal is not fully constrained. For \Object{R} and \Object{P} considered as
symmetric objects, there exists a manifold of possible locations where the
pouring is feasible to execute (as long as \Object{P} is located above \Object{R}, keeping its top surface above the opening of \Object{R}, the tilting is likely to succeed). Given this, we also have to generate possible goal grasp poses 
(that allow tilting) in order to evaluate our metric in a similar manner 
as in the pick-and-place case.

\subsection{Generation of possible grasp poses}

\begin{figure}[h]
\centering
\begin{tikzpicture}[framed,background rectangle/.style={thick,draw=black,fill=bgcolor,rounded corners=1em}]
\node[name=genPosesA]{\includegraphics[width=0.6\linewidth]{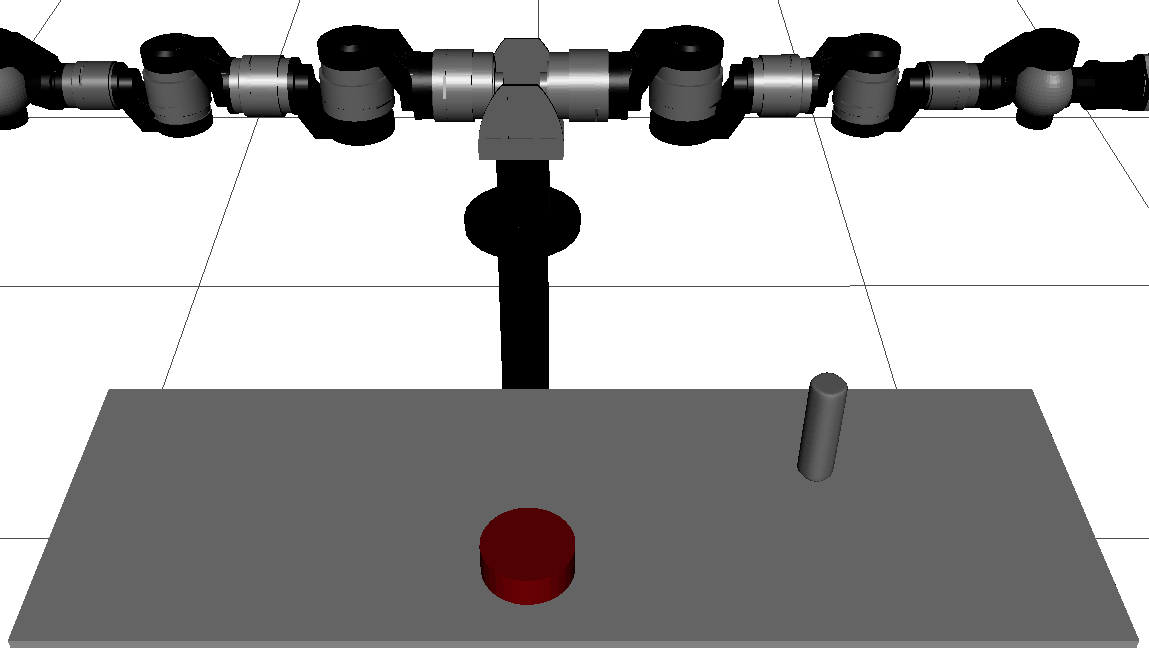}};
\node[name=genPosesB,yshift=-2.25cm]{\includegraphics[width=0.9\linewidth]{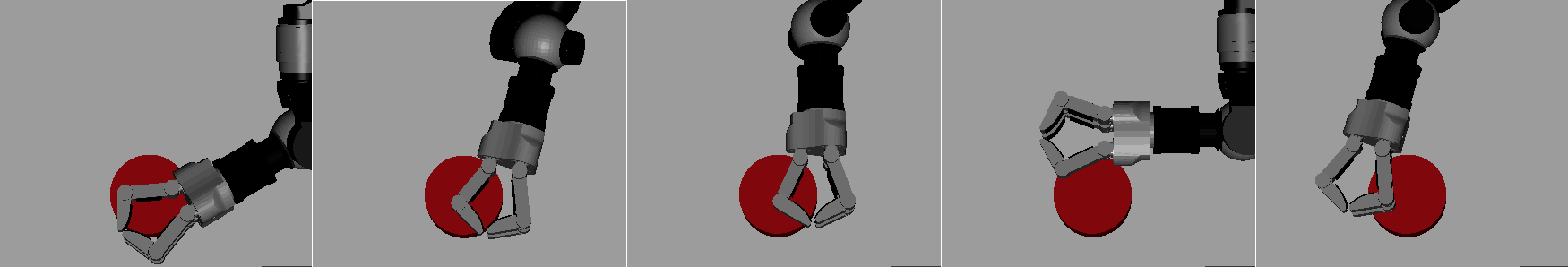}};
\end{tikzpicture}
\caption{Generating a set of discretized guess goal poses around the container \Object{R} . Notice in this example that
most of the grasps come from ``behind'' \Object{R}, meaning that due to the \Object{R}~location on the table, the robot
can only reach from these positions. }
\label{fig:genGoalPoses_pouring}
\vspace{-2ex}
\end{figure}

Given a receiver \Object{R}, many possible hand poses near it can be considered as candidate goal poses
(nearest the \Object{R}'s center, nearest its border). We generate a set of discretized goal poses
that allow the hand to finish in an orientation suitable for pouring from symmetrical objects by setting the hand orientation such that its approach direction is tangent to \Object{R}'s perimeter and has its $z$ orientation up. The position of the hand is set as the average between the minimum
and maximum distance the object \Object{P}~ can be from \Object{R}~such that \Object{P}~perimeter is
on top of \Object{R}. \fref{fig:genGoalPoses_pouring} shows example goal guess poses generated
for a given \Object{R}~location. Some parts of the \Object{P}'s border are unreachable for the hand,
hence no grasps will be considered in that area. The formal algorithm of this process is shown
in Algorithm \ref{alg:genGuessPoses}. An example of the end-effector generated for a sample scenario
can be seen in \fref{fig:genGoalPoses_pouring}

\begin{algorithm}
\DontPrintSemicolon
\KwIn{\Tf{w}{s}, \Object{P}, \Object{R}}
\KwOut{\GoalSet{}: A set of end-effector configurations circumventing \Object{R}}
\SetKwFunction{altura}{.height}
\SetKwFunction{radius}{.radius}
\SetKwFunction{trans}{.trans}
\SetKwFunction{rot}{.rot}
\SetKwFunction{setPose}{set\_Pose}
\SetKwFunction{existIKSol}{exist\_IK\_sol}
\SetKw{true}{true}
\BlankLine
\tcc{Generate poses such that hand is tangent to the borders of \Object{R}}
\ForEach{ i $\in$ ${[0,N>}$} {
  $\theta \leftarrow (\dfrac{2\pi}{N})i$\;
\tcc{\radius{} is obtained from the SQ parameters describing \Object{R}~and \Object{P}}
  $r_{m} \leftarrow$ $\dfrac{1}{2}$\Object{R}\radius{} + \Object{P}\radius{} \;
  $\mathbf{p} \leftarrow {}^{w}P_{C} + (r_{m}\cos{\theta}, r_{m}\sin{\theta},$ \Object\altura{} + $\delta_{m})$ \;
  $\mathbf{x} \leftarrow (-\cos{\theta}, -\sin{\theta},0)$ \;
  $\mathbf{z} \leftarrow (\sin{\theta}, -\cos{\theta}, 0)$ \;
  $\mathbf{y} \leftarrow \mathbf{z} \times \mathbf{x} $ \;
  $R \leftarrow [\mathbf{x},\mathbf{y},\mathbf{z}]$ \;
  ${}^{w}T_{p}$\trans{} $\leftarrow$ $\mathbf{p}$ \;
  ${}^{w}T_{p}$\rot{} $\leftarrow$ $R$ \;
  \tcc{Check if an IK solution is available}
  \If{ \existIKSol{${}^{w}T_{p}$} \bf{is} \true }{
    \GoalSet{}\pushback{${}^{w}T_{p}$}\;
  }
}
\Return \GoalSet{}\;
\caption{generate\_goal\_guesses}
\label{alg:genGuessPoses}
\end{algorithm}
%\vspace{-5ex}

%%%%%%%%%%%%%%%%%%%%%%%%%%%%%%%%%%%%%%%%%%%%%%%%%%%%%%%%%%%%%%%%%
\subsection{Evaluation of $m_{ag}$ at start, goal and as an average}
\begin{figure}[h]
\centering
\begin{tikzpicture}[framed,background rectangle/.style={thick,draw=black,fill=bgcolor,rounded corners=1em}]
\node[name=setup]{\includegraphics[width=0.45\linewidth]{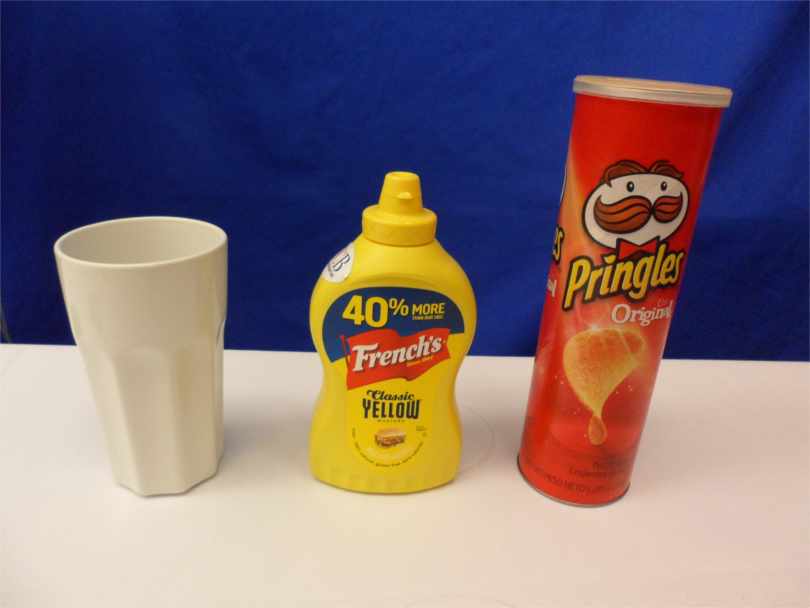}};
\node[name=setup,xshift=4cm]{\includegraphics[width=0.45\linewidth]{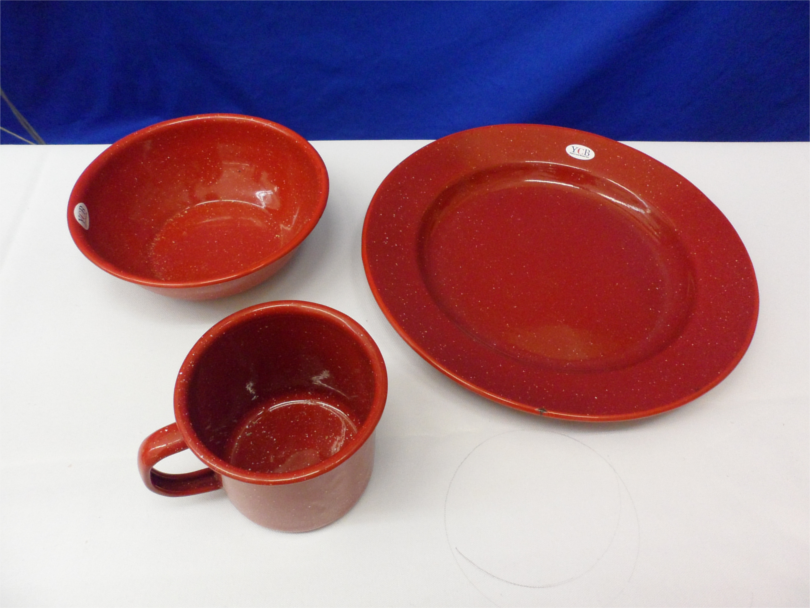}};
\end{tikzpicture}
\caption{Objects used in our physical experiments. Left: Objects being poured. Right: Containers set on table.}
\label{fig:pouring_objects_experiments}
\end{figure}

In a manner similar to Section \ref{subsec:EvalPaP}, we evaluate the
metric $m_{ag}$ measured at the start pose, guess goal pose and as an
average of both. For these pouring tasks, however, since there
are two objects involved (\Object{R}~and~\Object{P}), we test different
combinations of these. The object models used for these are shown in
\fref{fig:pouring_objects_experiments}

\begin{figure}[h]
\centering
\begin{tikzpicture}[framed,background rectangle/.style={thick,draw=black,fill=bgcolor,rounded corners=1em}]
\node[name=reach]{\includegraphics[width=0.95\linewidth]{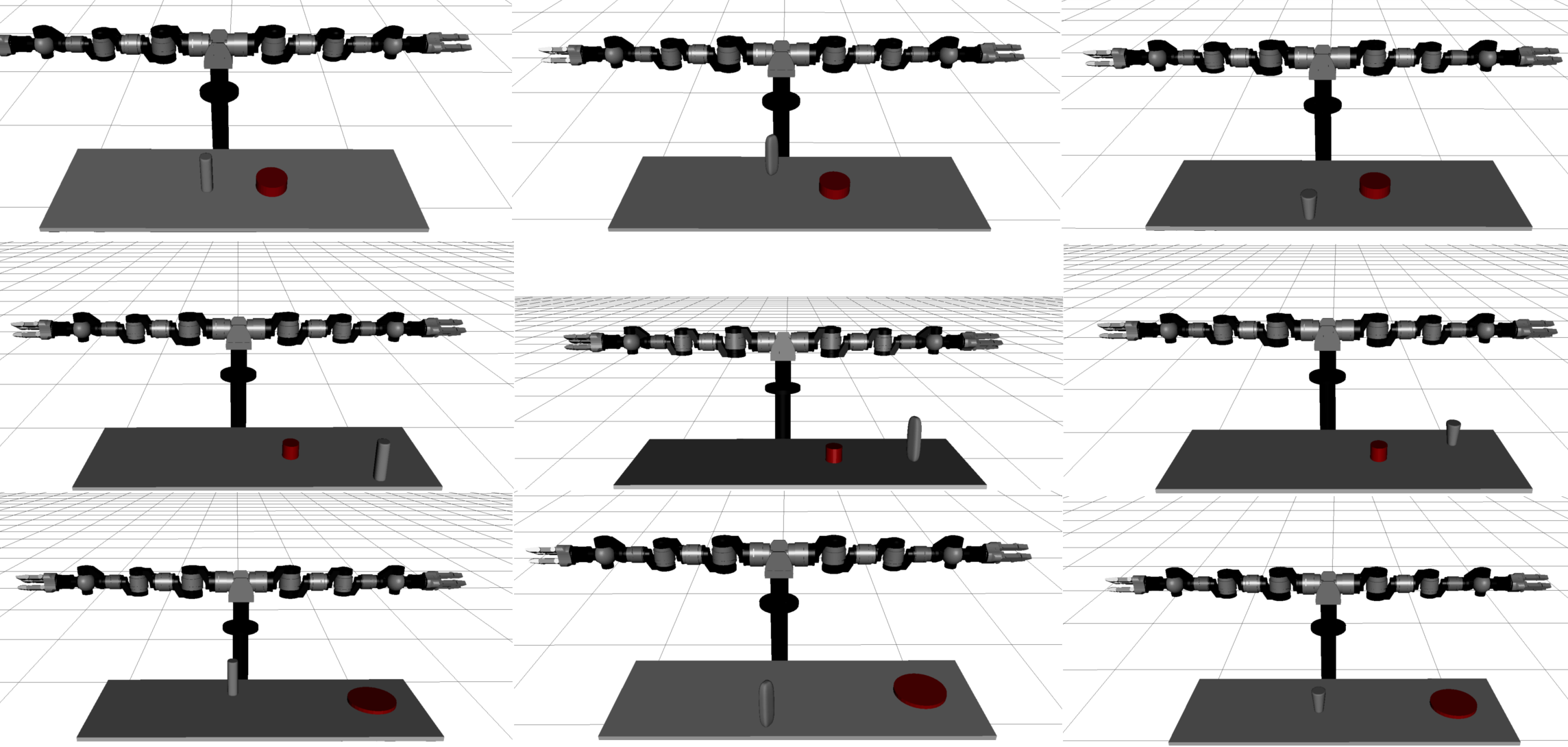}};
%\node[text=blue, xshift=0cm,yshift=-2cm]{Reaching};
\end{tikzpicture}
\caption{Sample simulation scenarios for pouring tasks combining the
receiver and pouring object randomly.}
\label{fig:pouring_simTestCases}
\vspace{-4ex}
\end{figure}

The results are shown in Tables \ref{tab:pouring_simResults_redCup}, \ref{tab:pouring_simResults_redPlate} and \ref{tab:pouring_simResults_redBowl}. In a surprising turn of events, for the pouring
tasks evaluated we found that the metric with the best performance both in success rate and end-effector displacement was $m_{ag}$ evaluated at the start location. This partly contradicts our findings
in the previous pick-and-place scenario, where the measure $m_{ag}$ taken using the average 
configurations between start and goal provided the best performance. Nonetheless, as in the previous
cases, we observe that the performance with metrics (any metric) is in general better than when
using any other grasp selected. We hypothetize that the main reason why in this case the start pose 
is more determinant in the results is that the goal pose for the end-effector really does not 
directly depend on the start pose. Rather,  independently of where the hand starts, the goal pose is mostly defined by the position of the container in the robot workspace (the hands ends up in a pose with respect to the container that is more
comfortable). Given this, it is not surprising then that the best performance is given only by the start location.

\begin{table}[h]
\scriptsize
\setlength{\tabcolsep}{3pt}
\centering
\caption{Simulation results of randomized scenarios (Container: Red cup)}
\label{tab:pouring_simResults_redCup}
\begin{tabular}{l |c c c|c c c|c c c}
\topline
\headcol \textbf{Object} & \multicolumn{3}{c|}{\textbf{Success}} & \multicolumn{3}{c}{\textbf{Hand Disp.(m)}} &  \multicolumn{3}{c}{\textbf{Plan time (s)}} \\
\headcol  & \textbf{Goal} & \textbf{Start} & \textbf{Avg} & \textbf{Goal} & \textbf{Start} & \textbf{Avg} & \textbf{Goal} & \textbf{Start} & \textbf{Avg} \\
\midline
\multirow{1}{*}{Pringles} & 221/250 & {\color{blue}250/250} & 240/250 & 2.20 & {\color{blue}2.14} & 2.18 & 1.86 & {\color{blue}1.10} & 1.37 \\
 %& 187/250 & 184/250 & 221/250 & 2.37 & 2.44 & 2.37 & 1.42 & 3.63 & 2.25 \\
\cmidrule{1-10} 
\multirow{1}{*}{White cup} & 230/250 & {\color{blue}250/250} & 237/250 & 2.18 & {\color{blue}2.10} & 2.13 & 1.71 & {\color{blue}1.17} & 1.37  \\
 %& 204/250 & 205/250 & 218/250 & 2.24 & 2.30 & 2.29 & 1.38 & 2.04 & 1.96 \\
\cmidrule{1-10}
\multirow{1}{*}{Soft Scrub} & 234/250 & {\color{blue}250/250} & 243/250 & 2.22 & {\color{blue}2.10} & 2.16 & 1.89 & {\color{blue}1.07} & 1.31  \\
 %& 200/250 & 200/250 & 228/250 & 2.43 & 2.42 & 2.42 & 1.83 & 2.94 & 2.09 \\
\bottomrule
\end{tabular}
\vspace{-4ex}
\end{table}

\begin{table}[h]
\scriptsize
\setlength{\tabcolsep}{3pt}
\centering
\caption{Simulation results of randomized scenarios (Container: Red plate)}
\label{tab:pouring_simResults_redPlate}
\begin{tabular}{l |c c c|c c c|c c c}
\topline
\headcol \textbf{Object} & \multicolumn{3}{c|}{\textbf{Success}} & \multicolumn{3}{c}{\textbf{Hand Disp.(m)}} &  \multicolumn{3}{c}{\textbf{Plan time (s)}} \\
\headcol  & \textbf{Goal} & \textbf{Start} & \textbf{Avg} & \textbf{Goal} & \textbf{Start} & \textbf{Avg} & \textbf{Goal} & \textbf{Start} & \textbf{Avg} \\
\midline
\multirow{1}{*}{Pringles} & 220/250 & {\color{blue}250/250} & 242/250 & 2.20 & {\color{blue}2.10} & 2.14 & 2.35 & {\color{blue}1.08} & 1.40 \\
 %& 184/250 & 183/250 & 214/250 & 2.38 & 2.47 & 2.41 & 1.44 & 3.51 & 2.36 \\
\cmidrule{1-10} 
\multirow{1}{*}{White cup} & 236/250 & {\color{blue}250/250} & 240/250 & 2.18 & {\color{blue}2.09} & 2.12 & 1.67 & {\color{blue}1.07} & 1.32  \\
 %& 215/250 & 213/250 & 229/250 & 2.21 & 2.28 & 2.24 & 1.31 & 2.11 & 1.80 \\
\cmidrule{1-10}
\multirow{1}{*}{Soft Scrub} & 225/250 & {\color{blue}250/250} & 236/250 & 2.22 & {\color{blue}2.13} & 2.17 & 2.04 & {\color{blue}1.01} & 1.17  \\
 %& 206/250 & 206/250 & 229/250 & 2.38 & 2.40 & 2.37 & 1.49 & 2.69 & 1.71 \\
\bottomrule
\end{tabular}
\vspace{0.0em}
\end{table}

\begin{table}[h]
\scriptsize
\setlength{\tabcolsep}{3pt}

\centering
\caption{Simulation results of randomized scenarios (Container: Red bowl)}
\label{tab:pouring_simResults_redBowl}
\begin{tabular}{l |c c c|c c c|c c c}
\topline
\headcol \textbf{Object} & \multicolumn{3}{c|}{\textbf{Success}} & \multicolumn{3}{c}{\textbf{Hand Disp.(m)}} &  \multicolumn{3}{c}{\textbf{Plan time (s)}} \\
\headcol  & \textbf{Goal} & \textbf{Start} & \textbf{Avg} & \textbf{Goal} & \textbf{Start} & \textbf{Avg} & \textbf{Goal} & \textbf{Start} & \textbf{Avg} \\
\midline
\multirow{1}{*}{Pringles} & 222/250 & {\color{blue}249/250} & 240/250 & 2.20 & {\color{blue}2.11} & 2.13 & 2.56 & {\color{blue}1.30} & 1.80 \\
% & 182/250 & 183/250 & 207/250 & 2.39 & 2.44 & 2.42  & 1.65 & 4.06 & 2.66 \\
\cmidrule{1-10} 
\multirow{1}{*}{White cup} & 237/250 & {\color{blue}250/250} & 244/250 & 2.17 & {\color{blue}2.10} & 2.13 & 2.58 & {\color{blue}1.79} & 2.13  \\
% & 204/250 & 203/250 & 216/250 & 2.20 & 2.30 & 2.29 & 2.20 & 3.59 & 3.27 \\
\cmidrule{1-10}
\multirow{1}{*}{Soft Scrub} & 234/250 & {\color{blue}249/250} & 236/250 & 2.22 & {\color{blue}2.06} & 2.12 & 2.90 & {\color{blue}1.44} & 1.92  \\
% & 206/250 & 204/250 & 227/250 & 2.40 & 2.42 & 2.43 & 2.37 & 3.90 & 2.91 \\
\bottomrule
\end{tabular}
\vspace{-4ex}
\end{table}

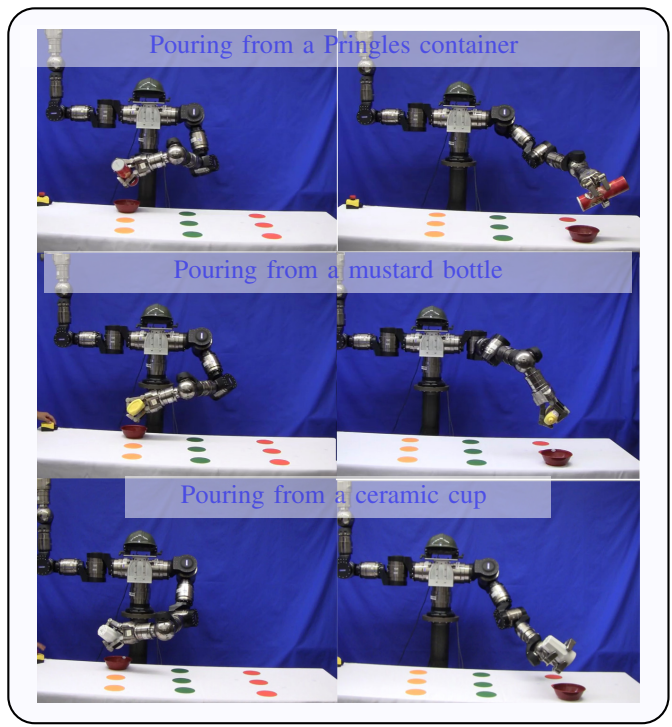
\begin{figure}[h]
\vspace{-2ex}
\centering
\begin{tikzpicture}[framed,background rectangle/.style={thick,draw=black,fill=bgcolor,rounded corners=1em}]
\node[]{\includegraphics[width=0.45\textwidth]{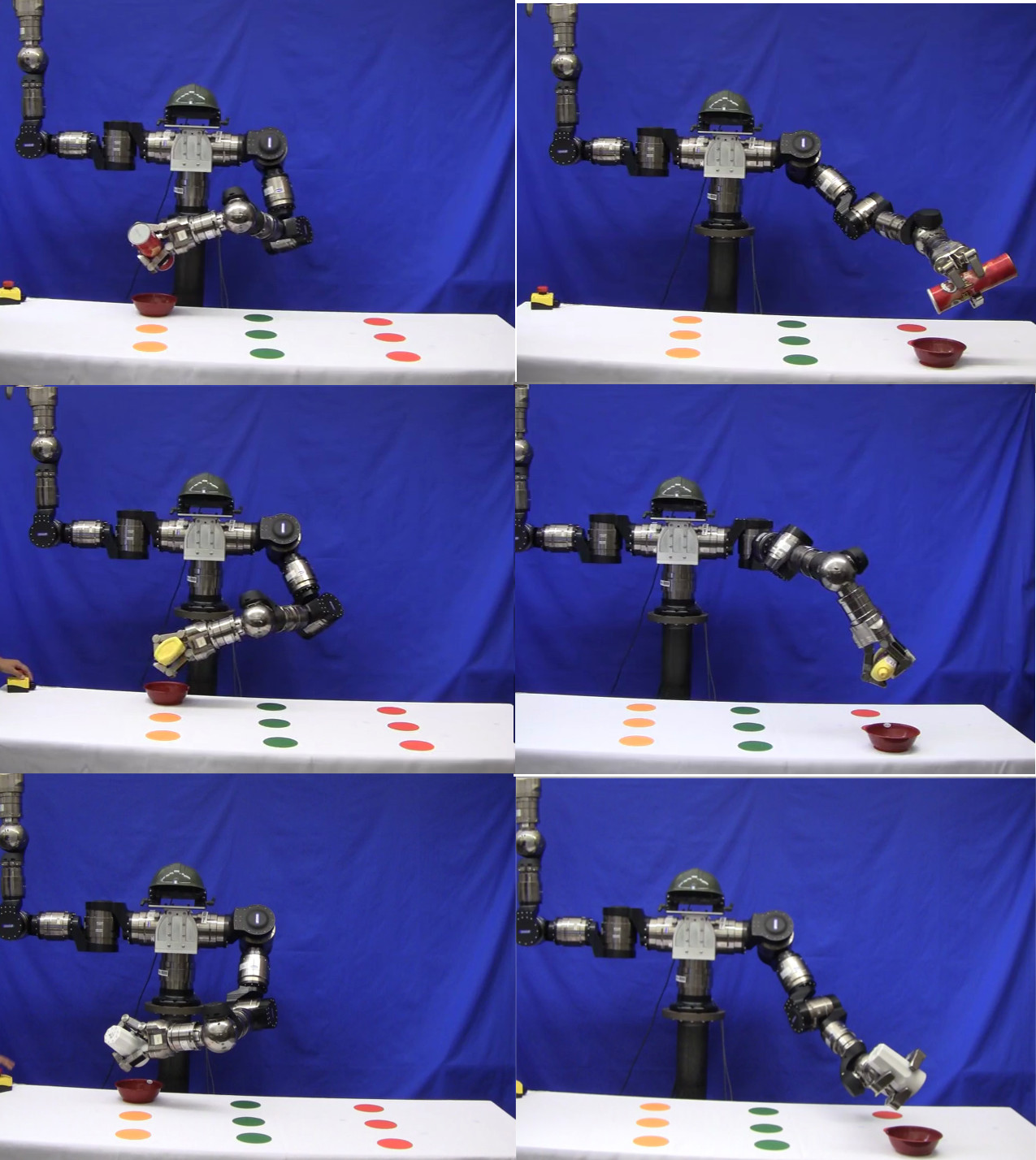}};
%\draw[line width=0.75mm, white] (-4,0) -- (4,0);
\node[text=blue,xshift=0cm,yshift=4.25cm,fill=white,fill opacity=0.5]{~~~~~~~~~~~~~Pouring from a Pringles container~~~~~~~~~~~~~~~};
%\node[text=blue,xshift=-2.0cm,yshift=1.75cm]{Container far bottom right};
%\node[text=blue,xshift=2.0cm,yshift=1.75cm]{Container far top left};
\node[text=blue,xshift=0cm,yshift=1.25cm,fill=white,fill opacity=0.5]{~~~~~~~~~~~~~Pouring from a mustard bottle~~~~~~~~~~~~~~};
%\node[text=blue,xshift=-2.0cm,yshift=-1.25cm]{Container far bottom right};
%\node[text=blue,xshift=2.0cm,yshift=-1.25cm]{Container far top left};
\node[text=blue,xshift=0cm,yshift=-1.75cm,fill=white,fill opacity=0.5]{~~~~~Pouring from a ceramic cup ~~~~~~};
%\node[text=blue,xshift=-2.0cm,yshift=-4.25cm]{Container far bottom right};
%\node[text=blue,xshift=2.0cm,yshift=-4.25cm]{Container far top left};
\end{tikzpicture}
\caption{Pouring: Final solutions for similar goal constraints are 
also similar, regardless of the start state (different for the  3 examples shown)}
\label{fig:pouring_3_examples}
\vspace{-2.0em}
\end{figure}

As for the pick-and-place case, the results showed here mostly involve simulation. We also tested
our grasp selection approach in our physical robot, performing 54 runs involving 3 containers, 3 
objects to pour from (with different geometry) and different start and goal locations. Some of these
results are shown in the accompanying video, and a few snapshots are shown in \fref{fig:pouring_3_examples},
which depicts the robot final goal state during pouring tasks at 2 locations. For each object, the start
location of the object was different, however we can see that the final state is similar. As we discussed earlier in this Section, the goal for the pouring tasks is loosely constrained in such a way that the goal 
state is not strongly tied to the start state, hence it does not matter too much for the grasp selection process.

%-------------------------
% PaP cheezit
%-------------------------

\section{Conclusions}
\label{sec:Conclusions}
In this paper we have presented a quantitative analysis of the advantages of using a manipulation metric ($m_{ag}$) to select a grasp from a set of possible candidates. By using our metric, we observed that the grasp selected in most of the cases entails arm motions that present 3 advantageous characteristics: (1) Shorter end-effector path lengths, (2) Shorter planning times, and (3) Higher success rate with respect to other grasps in the candidate set. We evaluated our metric under 3 different modalities: At the start step, at the goal step and as an average of both. We found that for pick-and-place tasks, in which the goal constraints are more limited, the $m_{ag}$ measured as an average (that is, considering both the start and the goal states) produced better results. For pouring tasks, in which the goal state is even more loosely defined, we found that by simply measuring our metric at the start step, the results were better when comparing it with any of the other grasps in the candidate set. 

As it was clearly stated in the abstract, this paper focuses on two-step tasks. Even in this simple case we found that there is not a single one-size-fits-all strategy to select adequate grasps that present comparative better properties as the ones discussed in this paper (brief planning times, end-effector short paths, high success rate). The approach presented here is not directly transferable to 3-or-more-step tasks. A possible way to circumvent this issue could be dividing complex manipulation tasks in simple one- and two-step tasks (in which our metric could be used to select the adequate grasps). This is matter of future work.

%%%%%%%%%%%%%%%%%%%%%%%%%%%%%%%%%%%%%%%%%
%% Bibliography
%%%%%%%%%%%%%%%%%%%%%%%%%%%%%%%%%%%%%%%%%
\bibliography{humanoids2016}

\begin{thebibliography}{10}

\bibitem{balasubramanian2014physical}
R.~Balasubramanian, L.~Xu, P.~D Brook, J.R. Smith, and Y.~Matsuoka.
\newblock Physical human interactive guidance: Identifying grasping principles
  from human-planned grasps.
\newblock In {\em The Human Hand as an Inspiration for Robot Hand Development}.
  Springer, 2014.

\bibitem{berenson2007grasp}
D.~Berenson, R.~Diankov, K.~Nishiwaki, S.~Kagami, and J.~Kuffner.
\newblock Grasp planning in complex scenes.
\newblock In {\em Humanoids}, 2007.

\bibitem{berenson2011task}
D.~Berenson, S.~Srinivasa, and J.~Kuffner.
\newblock Task space regions: A framework for pose-constrained manipulation
  planning.
\newblock {\em The International Journal of Robotics Research}, 2011.

\bibitem{bohg2014data}
J.~Bohg, A.~Morales, T.~Asfour, and D.~Kragic.
\newblock Data-driven grasp synthesis: A survey.
\newblock {\em IEEE Transactions on Robotics}, 2014.

\bibitem{cutkosky1989grasp}
M.R. Cutkosky.
\newblock On grasp choice, grasp models, and the design of hands for
  manufacturing tasks.
\newblock {\em IEEE Transactions on Robotics and Automation}, 1989.

\bibitem{ferrari1992planning}
C.~Ferrari and J.~Canny.
\newblock Planning optimal grasps.
\newblock In {\em ICRA}, 1992.

\bibitem{finn2016deep}
C.~Finn and S.~Levine.
\newblock Deep visual foresight for planning robot motion.
\newblock {\em arXiv preprint arXiv:1610.00696}, 2016.

\bibitem{roa2014integrated}
J.~Fontanals, B.A. Dang-Vu, O.~Porges, J.~Rosell, and M.~Roa.
\newblock Integrated grasp and motion planning using independent contact
  regions.
\newblock {\em Humanoids}, 2014.

\bibitem{gu2016deep}
S.~Gu, E.~Holly, T.~Lillicrap, and S.~Levine.
\newblock Deep reinforcement learning for robotic manipulation with
  asynchronous off-policy updates.
\newblock {\em arXiv preprint arXiv:1610.00633}, 2016.

\bibitem{huaman2016metric}
A.~Huam\'{a}n~Quispe, H.~Ben~Amor, and H.I. Christensen.
\newblock Combining arm and hand metrics for sensible grasp selection.
\newblock In {\em CASE}. IEEE, 2016.

\bibitem{huaman2015exploiting}
A.~Huam{\'a}n~Quispe, B.~Milville, M.~Guti{\'e}rrez, C.~Erdogan, M.~Stilman,
  H.I. Christensen, and H.~Ben~Amor.
\newblock Exploiting symmetries and extrusions for grasping household objects.
\newblock In {\em ICRA}, pages 3702--3708. IEEE, 2015.

\bibitem{kroemer2015towards}
O.~Kroemer, C.~Daniel, G.~Neumann, H.~van Hoof, and J.~Peters.
\newblock Towards learning hierarchical skills for multi-phase manipulation
  tasks.
\newblock In {\em ICRA}, 2015.

\bibitem{leon2014characterization}
B.~Le{\'o}n, C.~Rubert, J.~Sancho-Bru, and A.~Morales.
\newblock Characterization of grasp quality measures for evaluating robotic
  hands prehension.
\newblock In {\em ICRA}, pages 3688--3693. IEEE, 2014.

\bibitem{levine2016learning}
S.~Levine, P.~Pastor, A.~Krizhevsky, and D.~Quillen.
\newblock Learning hand-eye coordination for robotic grasping with deep
  learning and large-scale data collection.
\newblock {\em arXiv preprint arXiv:1603.02199}, 2016.

\bibitem{nagasaki1989asymmetric}
H~Nagasaki.
\newblock Asymmetric velocity and acceleration profiles of human arm movements.
\newblock {\em Experimental Brain Research}, 74(2):319--326, 1989.

\bibitem{przybylski2011human}
M.~Przybylski, T.~Asfour, R.~Dillmann, R.~Gilster, and H.~Deubel.
\newblock Human-inspired selection of grasp hypotheses for execution on a
  humanoid robot.
\newblock In {\em IEEE-RAS Humanoids}, 2011.

\bibitem{rosenbaum1996cognition}
D.~A. Rosenbaum, C.~M. van Heugten, and G.~E. Caldwell.
\newblock From cognition to biomechanics and back: The end-state comfort effect
  and the middle-is-faster effect.
\newblock {\em Acta psychologica}, 94(1):59--85, 1996.

\bibitem{sahbani2012overview}
A.~Sahbani, S.~El-Khoury, and P.~Bidaud.
\newblock An overview of 3d object grasp synthesis algorithms.
\newblock {\em Robotics and Autonomous Systems}, 60(3):326--336, 2012.

\bibitem{shimizu2008analytical}
M.~Shimizu, H.~Kakuya, W.~Yoon, K.~Kitagaki, and K.~Kosuge.
\newblock {Analytical inverse kinematic computation for 7-DOF redundant
  manipulators with joint limits and its application to redundancy resolution}.
\newblock {\em Transactions on Robotics}, 24(5):1131--1142, 2008.

\bibitem{vahrenkamp2012simultaneous}
N.~Vahrenkamp, T.~Asfour, and R.~Dillmann.
\newblock Simultaneous grasp and motion planning: Humanoid robot armar-iii.
\newblock {\em Robotics \& Automation Magazine}, 2012.

\bibitem{yamaguchi2015pouring}
A.~Yamaguchi, C.G. Atkeson, and T.~Ogasawara.
\newblock Pouring skills with planning and learning modeled from human
  demonstrations.
\newblock {\em International Journal of Humanoid Robotics}, 12(03):1550030,
  2015.

\end{thebibliography}
\bibliographystyle{plain}

\end{document}